\documentclass[lettersize,journal]{IEEEtran}
\usepackage{amsmath,amsfonts}
\usepackage{algorithmic}
\usepackage{algorithm}
\usepackage{array}
\usepackage{textcomp}
\usepackage{stfloats}
\usepackage{url}
\usepackage{verbatim}
\usepackage{graphicx}
\usepackage{subcaption}
\usepackage{caption}
\usepackage{cite}
\usepackage{siunitx}
\usepackage{stanli}
\usepackage{ulem}
\usepackage{tabularx}
\usepackage{booktabs}
\usepackage{wrapfig}
\usepackage{afterpage}
\usepackage{svg}
\usepackage[accsupp]{axessibility}
\usepackage[absolute,overlay]{textpos}

\newcolumntype{d}{X}
\newcolumntype{s}{>{\hsize=.35\hsize}X}
\newif\ifarxiv

\arxivtrue

\usepackage[utf8]{inputenc}
\usepackage[T1]{fontenc}

\usepackage[inline]{enumitem}

\usepackage{hyperref}

\usepackage[capitalize]{cleveref}

\DeclareMathOperator*{\argmin}{arg\,min}

\usepackage{tikz}
\usetikzlibrary{shapes,arrows}
\usetikzlibrary{automata, arrows.meta, positioning}

\newcommand{\com}[0]{CoM}
\newcommand{\Geranos}[0]{{Geranos}}
\newcommand{\foldangle}[0]{\alpha}
\newcommand{\frictioncoeff}[0]{\mu_s}
\newcommand{\poleradius}[0]{r}
\newcommand{\inradius}[0]{R}
\newcommand{\F}[0]{\mathcal{F}}
\newcommand{\f}[0]{\boldsymbol{f}}

\newcommand{\mAh}{\milli\ampere{}\hour}
\DeclareSIUnit[number-unit-product = {}]{\inchQ}{\text{\textquotedbl}}
\DeclareSIUnit[number-unit-product = {}]{\rpm}{RPM}
\DeclareSIUnit{\KV}{KV}

\newcommand{\review}[1]{{#1}}
\newcommand{\strike}[1]{}

\begin{document}

\ifarxiv
\begin{textblock*}{\textwidth}(20mm,8mm) 
\noindent
\small \copyright 2023 IEEE. Personal use of this material is permitted. Permission from IEEE must be obtained for all other uses, in any current or future media, including reprinting/republishing this material for advertising or promotional purposes, creating new collective works, for resale or redistribution to servers or lists, or reuse of any copyrighted component of this work in other works.
\end{textblock*}
\fi

\title{Geranos: a Novel Tilted-Rotors Aerial Robot\\ for the Transportation of Poles}

\author{N. Gorlo*; S. Bamert*; R. Cathomen*; G. K\"appeli*; M. M\"uller*; T. Reinhart*; \\ H. Stadler*; H. Shen; E. Cuniato; M. Tognon; R. Siegwart
\thanks{*: co-first, equal contribution}}


\maketitle

\afterpage{%
\begin{figure}[t!]
    \centering
    \ifarxiv
    \includegraphics[width = 0.9\linewidth]{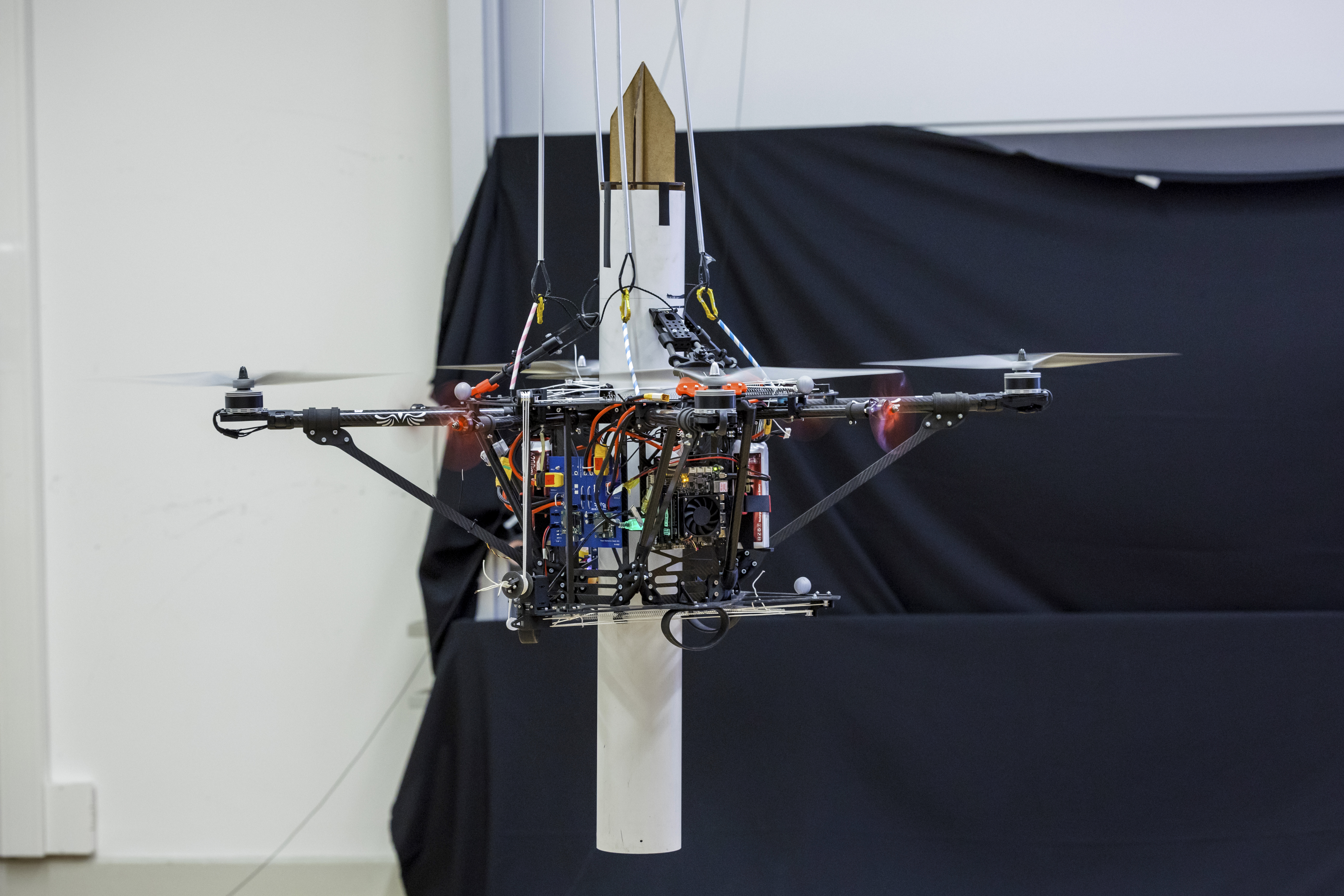}
    \else
    \includegraphics[width = 0.9\linewidth]{images/submission/compressed/1_Boreas_with_pole.jpg}
    \fi
    \caption{\review{The \Geranos{} platform carrying a pole.}}
    \label{fig:plattform}
\end{figure}
}

\begin{abstract}
In challenging terrains, constructing  structures such as antennas and cable-car masts often requires the use of helicopters to transport loads via ropes. 
The swinging of the load, exacerbated by wind, impairs positioning accuracy, therefore necessitating precise manual placement by ground crews.
This increases costs and risk of injuries.
Challenging this paradigm, we present \textit{\Geranos{}}: a specialized multirotor Unmanned Aerial Vehicle (UAV) designed to enhance aerial transportation and assembly. \Geranos{} demonstrates exceptional prowess in accurately positioning vertical poles, achieving this through an innovative integration of load transport and precision. Its unique  ring design mitigates the impact of high pole inertia, while a lightweight two-part grasping mechanism ensures secure load attachment without active force. With four primary propellers countering gravity and four auxiliary ones enhancing lateral precision, \Geranos{} achieves comprehensive position and attitude control around hovering. Our experimental demonstration mimicking antenna/cable-car mast installations showcases \Geranos{}' ability in stacking poles (3\,kg, 2\,m long) with remarkable sub-5\,cm placement accuracy, without the need of human manual intervention.

\end{abstract}


\section{Introduction}
\label{sec:intro}
\vspace{-5pt}

\begin{figure*}[t]

\begin{tabular}[c]{@{}c@{}}
    \begin{subfigure}[c]{.33\linewidth}
        \centering
        \ifarxiv
        \includegraphics[height=1.2\linewidth]{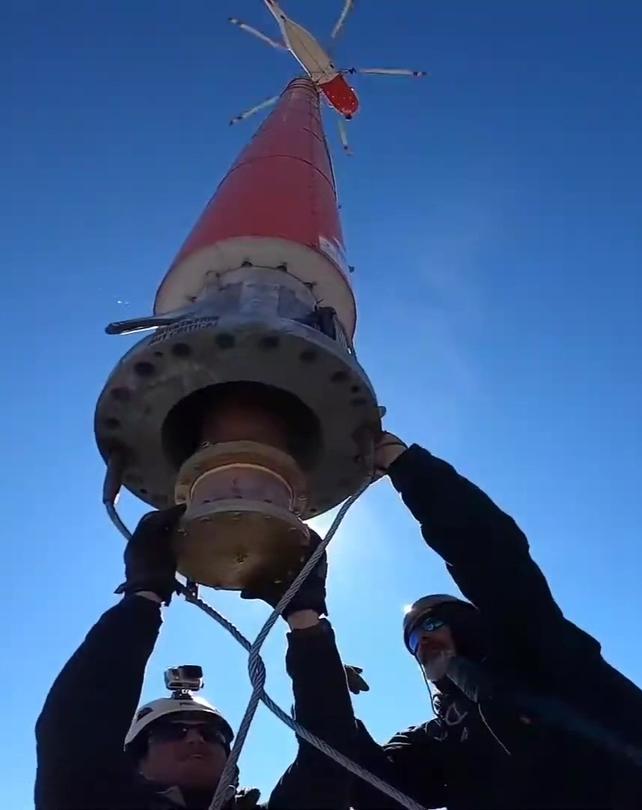}%
        \else
        \includegraphics[height=1.2\linewidth]{images/submission/compressed/2a_sling_load.jpg}%
        \fi
        \caption{}
        \label{fig:Guiding_load}
    \end{subfigure}
    \hfill
    \begin{subfigure}[c]{.33\linewidth}
        \centering
        \ifarxiv
        \includegraphics[height=1.2\linewidth]{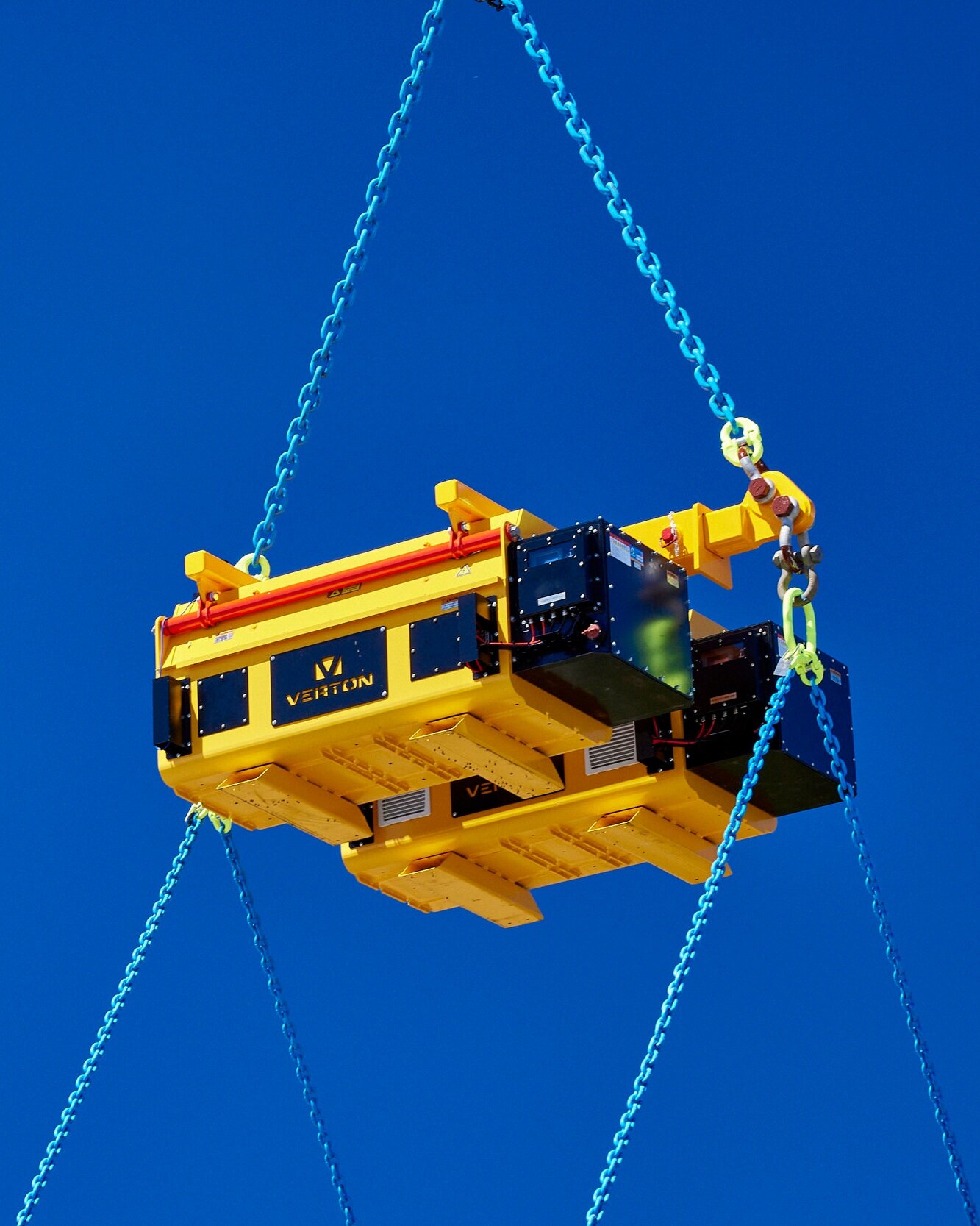}%
        \else
        \includegraphics[height=1.2\linewidth]{images/submission/compressed/2b_Verton2.jpg}%
        \fi
        \caption{}
        \label{fig:Verton}
    \end{subfigure}\\
    \noalign{\bigskip}%
\end{tabular}
\hfill
\begin{tabular}[c]{@{}c@{}}
    \begin{subfigure}[c]{.3\linewidth}
        \centering
        \ifarxiv
        \includegraphics[width=\linewidth]{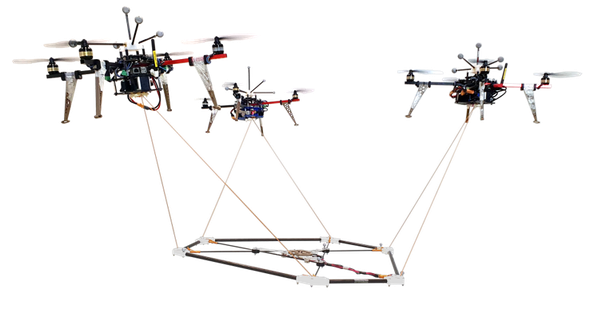}%
        \else
        \includegraphics[width=\linewidth]{images/submission/compressed/2c_flycrane.png}%
        \fi
        \caption{}
        \label{fig:SlingloadCooperative}
    \end{subfigure}\\
    \noalign{\bigskip}%
    \begin{subfigure}[c]{.3\linewidth}
        \centering
        \ifarxiv
        \includegraphics[width=0.5\linewidth,page=2]{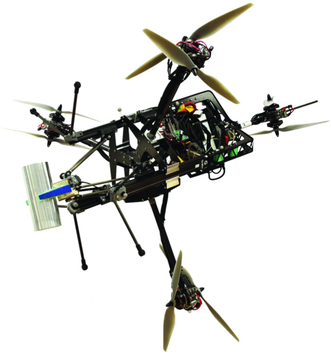}%
        \else 
        \includegraphics[width=0.5\linewidth,page=2]{images/submission/compressed/2d_prismav.png}%
        \fi
        \caption{}
        \label{fig:PrisMAV}
    \end{subfigure}
\end{tabular}
 \caption{
 (a) Common sling load operation for load\protect\footnotemark[4]; (b) An example of a commercial solution for load stabilization by Verton\protect\footnotemark[2]; (c) Fly-Crane system for cooperative sling load transportation \protect\cite{jimenez2022precise}; (d) The PrisMAV omnidirectional aerial manipulator with  a linear delta arm \protect\cite{PrisMAV}}
\label{fig:existing_solutions}
\end{figure*}

\IEEEPARstart{I}{n} contemporary construction sites, helicopters and cranes play vital roles in moving and erecting bulky loads. Yet, these loads are tethered to the vehicle using only ropes, which severely hinders accurate positioning. As a result, ground crews are necessary to achieve the needed precision, employing taglines attached to the load for directing it to the desired location (see \cref{fig:Guiding_load}).
Such operations are not without risk. For instance, crews can be endangered by a swinging load, especially when working in vulnerable areas like atop antennas, leading to numerous injuries and fatalities. While the German Social Accident Insurance reported over 3'300 accidents connected to cranes and their transported weight in 2020, the U.S. Bureau of Labor Statistics reported an average of 42 crane-related deaths per year throughout the period of 2011-2017. Furthermore, the European Union Aviation Safety Agency reported 10 helicopter accidents regarding construction in Europe in the year 2020, making construction and sling load operations the most dangerous special-helicopter-operation~\footnote{\review{Statistics acquired from "Arbeitsunfallgeschehen 2020. Deutsche Gesetzliche Unfallversicherung" (\url{https://publikationen.dguv.de/widgets/pdf/download/article/4271} , pg. $78$, accessed 2023-09-02), "Fatal Occupational Injuries Involving Cranes" (\url{https://www.bls.gov/iif/oshwc/cfoi/cranes-2017.htm}, accessed: 2022-09-10), and "Annual safety review 2021. European Union Aviation Safety Agency" (\url{https://www.easa.europa.eu/en/document-library/general-publications/annual-safety-review-2021}, pg. $106$, accessed 2023-09-02)}}.
Among those, this work focuses on the transportation and assembly of long poles which is particularly dangerous and challenging from a control perspective because of their high moment of inertia.

To boost safety during such operations, companies have introduced solutions that partially control the swinging load.
Examples are the \review{Vita load navigator}, Verton EVEREST Series and Roborigger AR15.\footnote{“Vita load navigator” (\url{https://www.vitaindustrial.co/vita-load-navigator}, accessed: 2022-09-02), “Verton EVEREST Series” (\url{https://www.verton.com.au/everest-series}, accessed: 2022-09-10), and “Roborigger AR15”, (\url{https://www.roborigger.com.au/products/roborigger-load-orientation-system/}, accessed: 2022-09-10)}
They are all based on the idea of attaching a system between the rope and the load (see \cref{fig:Verton}) capable \review{of orienting it by} exerting forces and torques \review{(via a thrust vectoring mechanism or gyroscope technology)}. 
However, due to their inability to control the loads' translational dynamics, they cannot provide the high precision that is often required for assembly tasks.

\review{The recent advancement in UAV autonomy has sparked interest in using UAVs for cargo transfer.} Titan\footnote{“Kaman titan”, \url{https://kaman.com/brands/kaman-air-vehicles/titan/}, accessed: 2022-09-02.}
\review{is a pilotless heavy lift helicopter that can carry over two tons but, because it uses a standard sling load technique for load transportation, it still suffers from swinging load and accuracy issues.}
\strike{a heavy lift helicopter without a pilot that can carry more than two tons, \strike{was unveiled by Kaman to show that human pilots are no longer necessary, even for huge payloads. The Titan} uses a standard sling load technique for load transportation and has a design that is essentially equivalent to a typical helicopter. As a result, the problems of a swinging load and a lack of accuracy still exist.}

\footnotetext[4]{"Antenna installation through Precision Communications" (\url{https://www.youtube.com/watch?v=NS7ZP7rGHR4}, accessed: 2022-10-31)}

In the last decade, researchers showed increasing interest in the challenging problem of load transportation with aerial robots~\cite{2021g-OllTogSuaLeeFra}. The early methods were based on UAVs like helicopters or quadcopters with various types of grippers attached to them. There are primarily two classes of grippers in use. 
The first method attaches the payload to the UAV, restricting movement to only the UAV's motion \cite{Yale}. This gripper prevents \review{load's swings} and improves precision. However, regular quadcopters and helicopters are underactuated, requiring tilting for movement. This makes precise object placement challenging, especially for objects with high inertia.
The second class of frequently used grippers is robotic arms~\cite{2021-BodTogSie}. 
AMUSE \cite{RoboArm} uses a 7 degrees of freedom arm which prevents the load from swinging and enables to reorient the load. However, due to the overactuation, the system becomes heavier and considerably more complicated to control.

Aiming at simpler systems, UAVs have also been employed with under-slung payloads. In \cite{uav_underslung_payload}, the authors show that the system is differentially-flat with respect to load position and yaw angle. However, the system has a relative degree of 6, meaning it can only compensate for external forces that are smooth up to the sixth degree. This leads to a delay in compensating for disturbances and interaction forces. As a result, previous works focused on transportation tasks and did not explore physical interaction tasks like pole assembly.

Multi-drone systems are another often proposed solution. 
The works in~\cite{jimenez2022precise,gabellieri2023equilibria,CooperativeManipulation, cooperative_transportation_Loianno} 
investigate sling load operations with multiple UAVs connected to the payload via cables \review{(see \cref{fig:SlingloadCooperative})}. While having the advantage of distributing the weight among multiple UAVs, such systems lead to more potential errors and the necessity for complicated coordination between UAVs. Additionally, when transporting a pole, cables would need to be attached to both the top and the bottom of the pole to exert the necessary torques controlling the pole's orientation.
This would further increase the system's complexity and proneness to errors.

All the proposed solutions so far suffer from different disadvantages: (i) the load is usually tethered to the vehicles, hindering its positioning accuracy; (ii) even if the system load is rigidly attached with a gripper, underactuation reduces accuracy and hinders the system's stability if the load's inertia is too big. 
To address these concerns, we identify three requirements four our aerial system: (i) the load needs to be rigidly attached to the vehicle; (ii) the load should be securely grasped in its center of mass (\com{}) such to reduce the effect of inertial disturbances; (iii) the vehicle should be fully actuated in order to translate without the necessity of tilting, thus increasing the maneuvering precision of the load.

Regarding the actuation, new multi-directional thrust systems have recently been introduced~\cite{2021-HamUsaSabStaTogFra}. These systems have been implemented in UAVs with tilted rotors \cite{hamandi2020omni}. Additionally, UAVs equipped with tiltable rotors have been introduced and paired with manipulators \cite{PrisMAV}, as depicted in ~\cref{fig:PrisMAV}. However, their potential for carrying bulky loads like poles, has not been fully exploited.

Regarding the grasping problem, several gripping mechanisms have been proposed for specific types of loads. A soft-robotic gripper design is presented in \cite{soft_robotic_gripper}, with three soft bending fingers and one passively adaptive palm. The authors in \cite{origami_gripper} present an origami-inspired reconfigurable suction gripper which can be used for cylindrical shapes. Due to the vertical orientation needed for poles, the existing grippers cannot grasp the \com{}, only the top end. This makes it more difficult to control the pole's position and orientation. Another potential solution is equipping a UAV with a spring-suspended manipulator, as described in \cite{Aerial_Manipulator_With_Elastic_Suspension}, providing accuracy, autonomy, and extensive payload capacity. However, this design's drawback is its inability to grip a pole's \com{}, again making pole orientation control challenging.

We therefore propose an entirely new solution: \textit{\Geranos{}}, the first UAV specifically designed for the aerial transportation and assembly of poles.
Our system combines three novel concepts concerning its structure, gripper and propeller configuration:
\begin{enumerate*}
    \item \review{To maximize \Geranos{}' control over the pole, the system is designed in a ring shape. This allows the pole to pass through the center hole, enabling \Geranos{} to fly over it and grip it at the \com{}, where the influence of the inertia is minimal.}
    \strike{In order to give \Geranos{} the best possible control over the pole, the system is built in a ring structure. This lets the pole pass through the center hole, which in turn allows \Geranos{} to fly over the pole and to grasp it at the \com{}, where the influence of the inertia is minimal.}  
    \item \review{Thanks to a specialized gripper design, \Geranos{} can easily center and grab the pole at its \com{} using only two actuators. Additionally, it can securely grip the pole without needing any support from the pole's structure. The centering mechanism is inspired on the Latching End Effector, which is part of the Space Station Remote Manipulator System \cite{nasa_centering}. While the Latching End Effector is built in a ring shape and is used for both centering and grabbing the payload, our mechanism is solely centering the payload and the strings are oriented such that they can be pulled by a single actuator.}
    \item \review{To place large poles accurately (small position error at its tip), \Geranos{} needs to fly upright. Small variations in tilt could cause major inaccuracies. To achieve this, we designed an octacopter with tilted rotors, allowing \Geranos{} to move sideways without tilting.} \review{In the current configuration of \Geranos{}, it can carry a weight of \qty{3}{\kg} for a flight duration of 10 minutes with a thrust to total weight ratio of \qty{1.61}{}:\qty{1}{}.} 
\end{enumerate*}

We demonstrate the \Geranos{} concept by autonomously transporting and stacking two vertical poles (up to \qty{3}{\kg}), showing how it could be used in real-world scenarios. As far as we know, this is the first rotary-wing VTOL UAV to perform such a task.

\section{System Layout} 
\label{sec:system_layout}

In this section, we discuss and justify the three features that makes \Geranos{} a unique system. 
\review{We first define two reference frames: the body frame $\F_b$, located at the geometric center of the UAV (not at its \com{}), and the world inertial frame $\F_w$, with the $z$ axis pointing up.} 
The $z$ axis of $\F_b$ points upwards, parallel to the thrust vectors of the main propellers. 

\subsection{Mechanical Structure}
\review{The mechanical structure is purpose-built for transporting} cylindrical poles, standing vertically at their initial and final position.
\strike{, i.e., the cylinder axis is parallel to the $z$-axis of $\F_w$.}
To minimize the inertial effects of the payload on the UAVs dynamics, the UAV \review{must} grab the payload such \review{to minimize} the distance between the \com{} of the payload and the UAV. As a result of this requirement, \review{we designed \Geranos{} with} a vertical clearance 
in the middle to accommodate the load, resulting in a ring like structure, as shown in~\cref{fig:geranos_sketch}.
This clearance was obtained by arranging six carbon panels in a hexagonal prism. The electronic components are mounted vertically on their outside. 
Two additional carbon plates are mounted on the top and the bottom of the hexagonal prism to ensure structural rigidity and to house the gripping mechanism.

\subsection{Gripper Design}
\label{sec:gripper}
Since grasping large poles with a UAV is a very specialized task, we designed an application-specific gripper. Our gripper has to be capable of grabbing cylindrical poles of different radii, length and weight. To be considered suitable for integration into a UAV, it has to be light-weight, energy-efficient, and accommodate for potential misalignments of the UAV relative to the pole. 
We achieved this by splitting the gripping mechanism in two subsystems, the \textit{centering mechanism} and the \textit{lifting mechanism}, each of which is actuated by one motor. The centering mechanism centers the pole in the middle of the hole\review{, which minimizes the distance in the body xy-plane between the \com{} of the pole and the one of the vehicle. This works by constraining the pole's degrees of freedom relative to the UAV by restricting translational movements in the $xy$-plane of $\F_b$ and rotation about any axis except for the $z$ axis}. The lifting mechanism prevents the pole from falling by using a self-locking clamping mechanism. \review{ It eliminates the remaining degrees of freedom of the pole by preventing it from moving and rotating along the $z$-axis of $F_b$.} 

\subsubsection{Centering}

\begin{figure*}
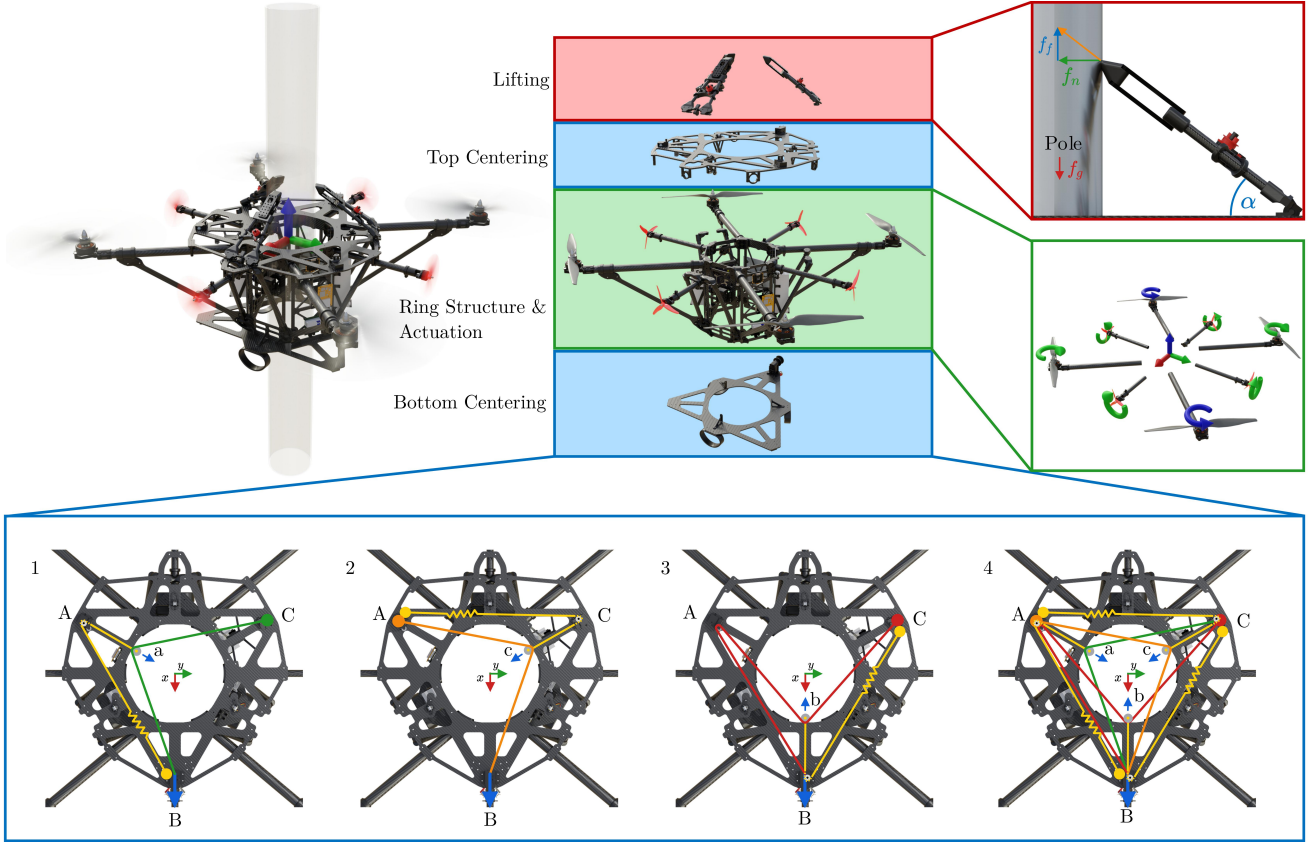

    \centering
    \ifarxiv
    \includegraphics[width=\linewidth]{images/submission/pdf/3_Geranos_Sketch.pdf}
    \else 
    \includegraphics[width=\linewidth]{images/submission/eps/3_Geranos_Sketch.eps}
    \fi
    \caption
      {\review{This figure illustrates the working principles of the \Geranos{} UAV. For greater clarity the vehicle is split into three parts: the lifting mechanism (red), the centering mechanism (blue) and the actuation (green). On the left, a picture of the full system is displayed, while in the middle an explosion view illustrates the three main parts of the UAV. The top right picture shows the relevant forces acting between the pole and one of the folding triangles used in the lifting mechanism. The picture in the middle right depicts the actuation setup and the spinning directions of all rotors (green for clockwise and blue for counter clockwise). The working principle of the centering mechanism is illustrated on the bottom of this figure. The yellow lines symbolize cables equipped with a spring while the green, orange and red lines represent inelastic cables. Cable fixations are represented by filled circles matching the color of the respective cables.
      }}
    \label{fig:geranos_sketch}
\end{figure*}
\review{In the blue box in \cref{fig:geranos_sketch}, the operating principle of the centering mechanism is depicted. The sketches 1, 2, and 3 depict parts of the mechanism, whereas sketch 4 provides a comprehensive view of its working principle. 
To constrain the pole in the middle, we want at least three contact points synchronously moving inward radially (outward to release the pole) with respect the center of the hole of the vehicle. Let us define $a$, $b$, and $c$ as the contact points, and $A$, $B$, and $C$ as the respective outermost end points (as shown in \cref{fig:geranos_sketch}). We will describe the centering mechanism for one contact point, $a$ (sketch 1). The mechanisms for the other two follow analogously (sketches 2 and 3). For the inward motion, we fix a cable (the green one) to the vertex $C$; the cable passes through $a$ via a pulley and then connects to the motor situated at $B$. When the motor pulls the cable, $a$ will move inward radially. Since $a$, $b$, and $c$ move synchronously, they will push the pole toward the center, constraining its position. Conversely, since $a$ is attached to $B$ via $A$ with a cable and a spring, when the motor releases the cable, the spring moves $a$ outward radially. All $a$, $b$, and $c$ perform the same motion, effectively releasing the pole.}

\begin{figure*}[t]
    \begin{subfigure}[t]{.32\linewidth}
        \ifarxiv
        \includegraphics[width=\linewidth]{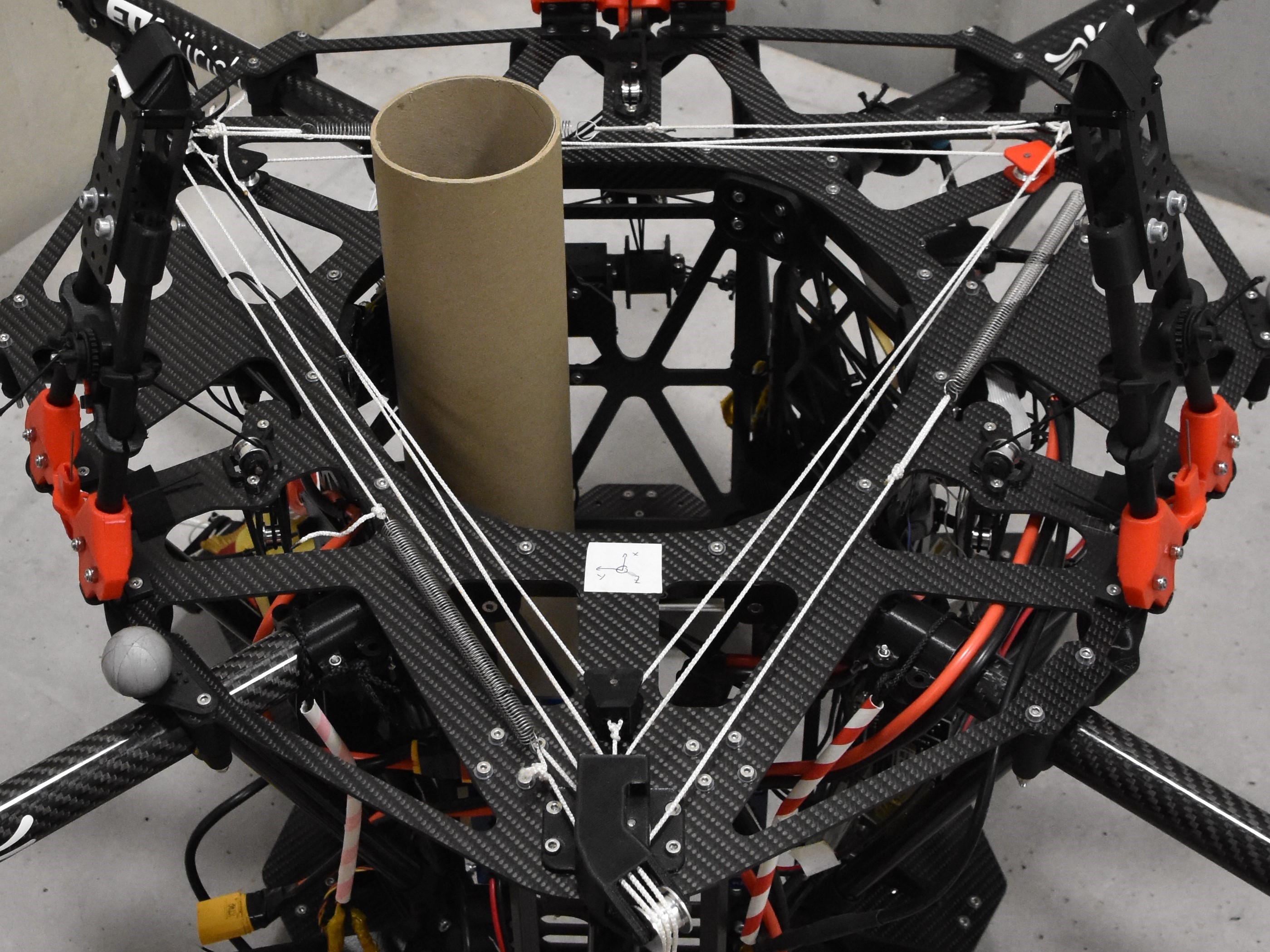}
        \else
        \includegraphics[width=\linewidth]{images/submission/high_quality/4a_centering_angle_1.jpg}
        \fi
        \caption{}
        \label{fig:gripping_full_a}
    \end{subfigure}
    \hfill
    \begin{subfigure}[t]{.32\linewidth}
        \ifarxiv
        \includegraphics[width=\linewidth]{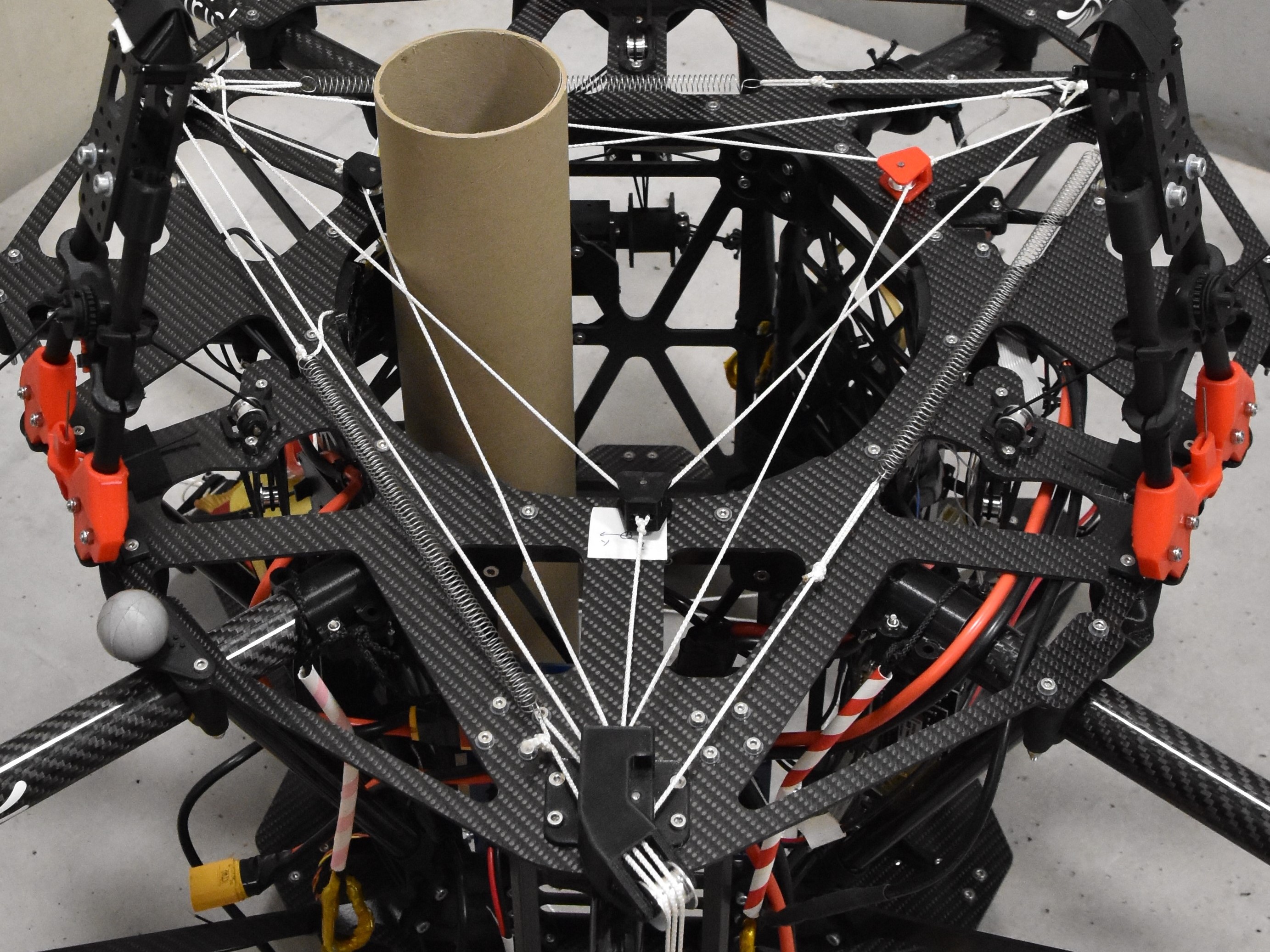}
        \else
        \includegraphics[width=\linewidth]{images/submission/high_quality/4b_centering_angle_2.jpg}
        \fi
        \caption{}
        \label{fig:gripping_full_b}
    \end{subfigure}
    \hfill
    \begin{subfigure}[t]{.32\linewidth}
        \ifarxiv
        \includegraphics[width=\linewidth]{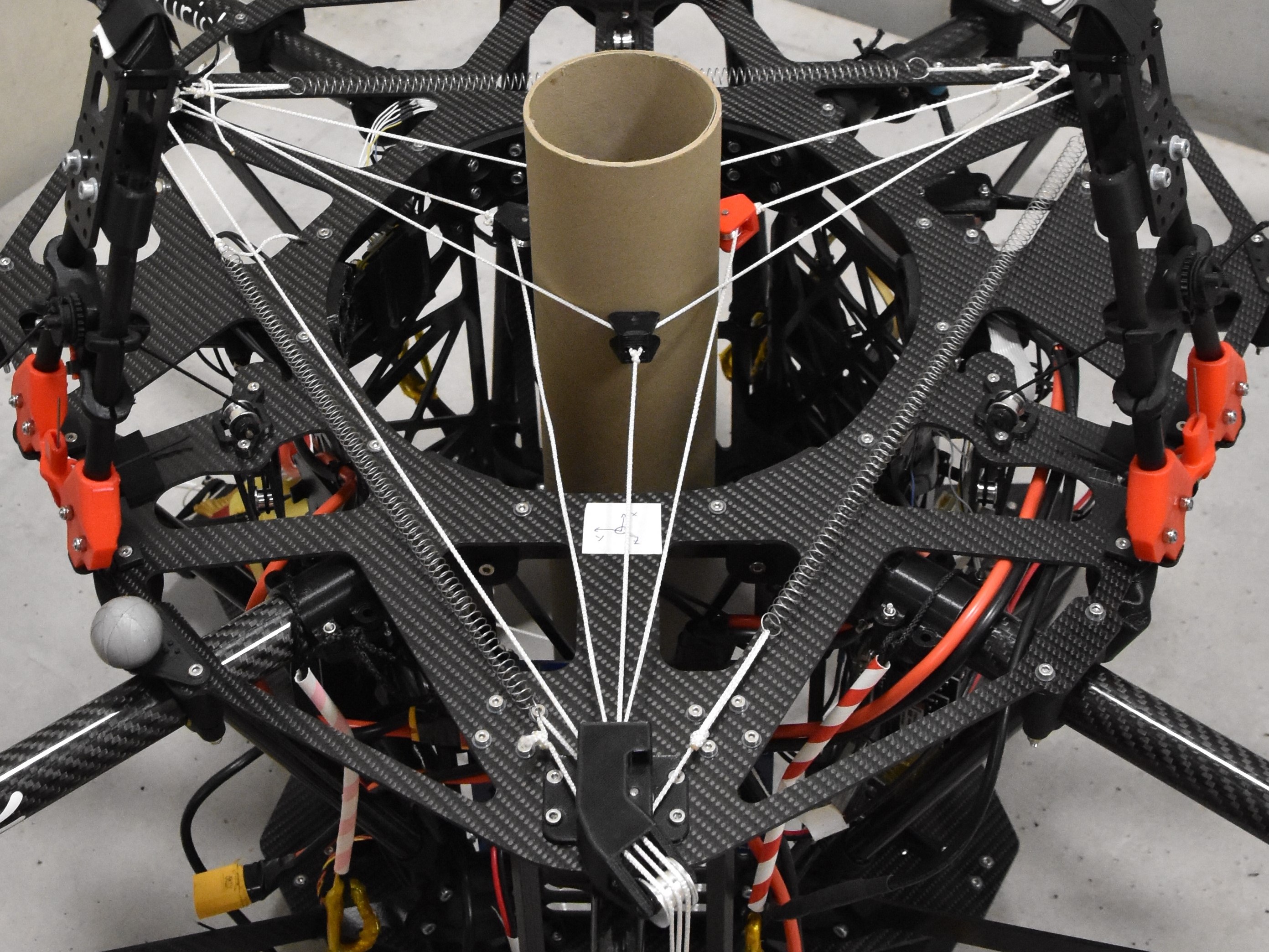}
        \else 
        \includegraphics[width=\linewidth]{images/submission/high_quality/4c_centering_angle_3.jpg}
        \fi
        \caption{}
        \label{fig:gripping_full_c}
    \end{subfigure}
    \vspace{.5cm}
    
    \begin{subfigure}[b]{.32\linewidth}
        \ifarxiv
        \includegraphics[width=\linewidth]{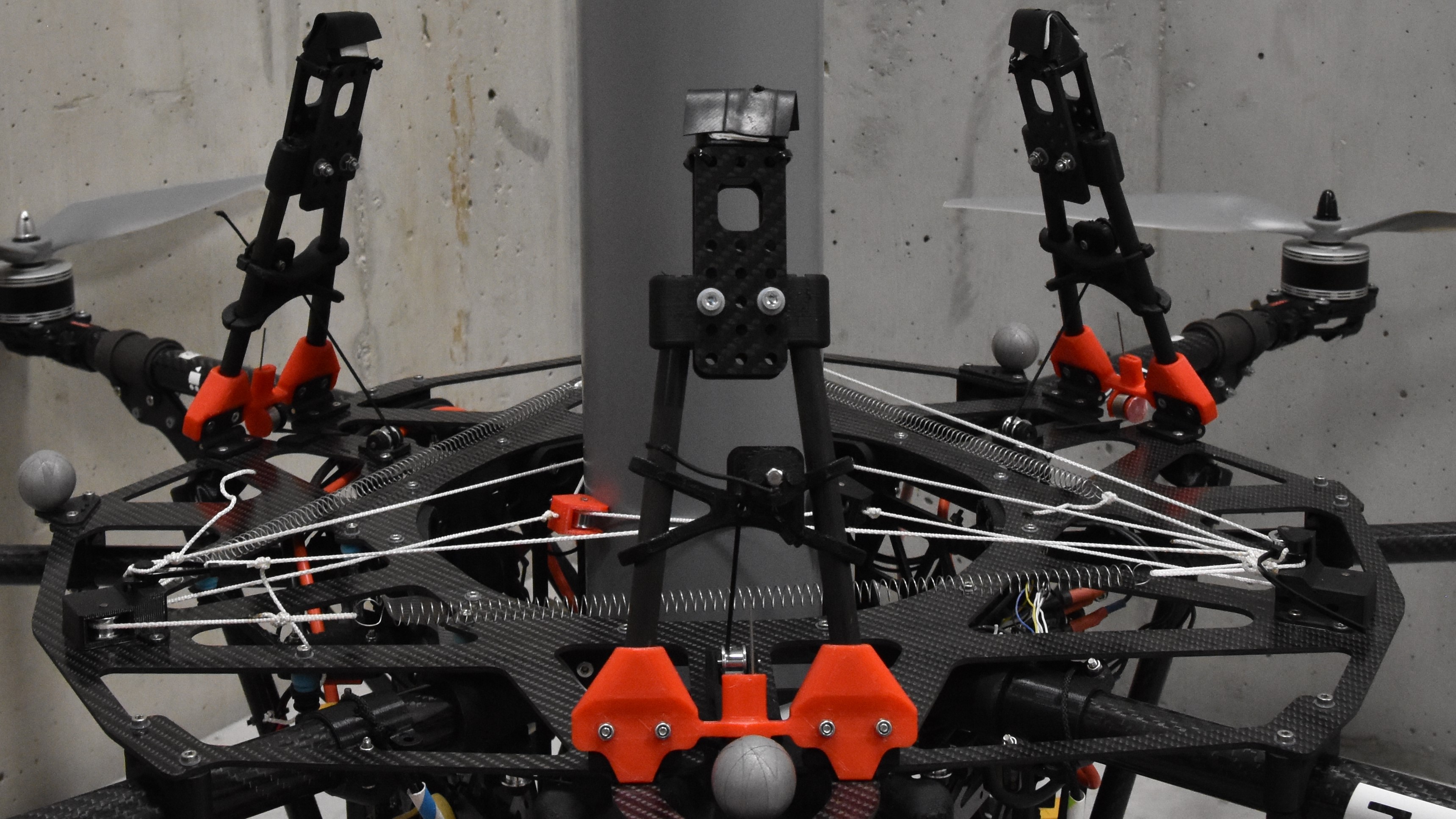}
        \else 
        \includegraphics[width=\linewidth]{images/submission/high_quality/4d_lift_1.jpg}
        \fi
        \caption{}
        \label{fig:gripping_full_d}
    \end{subfigure}
    \hfill
    \begin{subfigure}[b]{.32\linewidth}
        \ifarxiv
        \includegraphics[width=\linewidth]{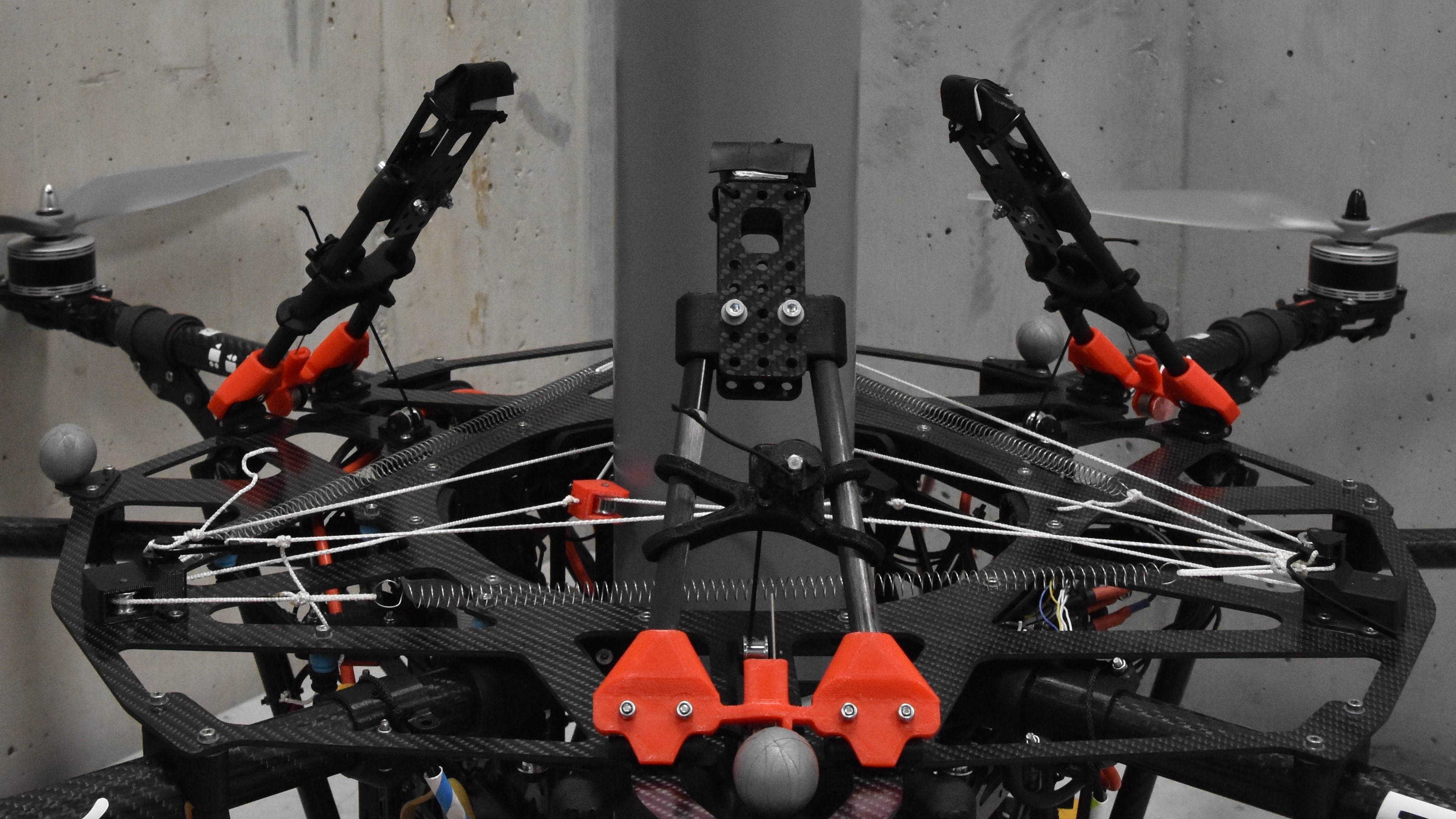}
        \else
        \includegraphics[width=\linewidth]{images/submission/high_quality/4e_lift_2.jpg}
        \fi
        \caption{}
        \label{fig:gripping_full_e}
    \end{subfigure}
    \hfill
    \begin{subfigure}[b]{.32\linewidth}
        \ifarxiv
        \includegraphics[width=\linewidth]{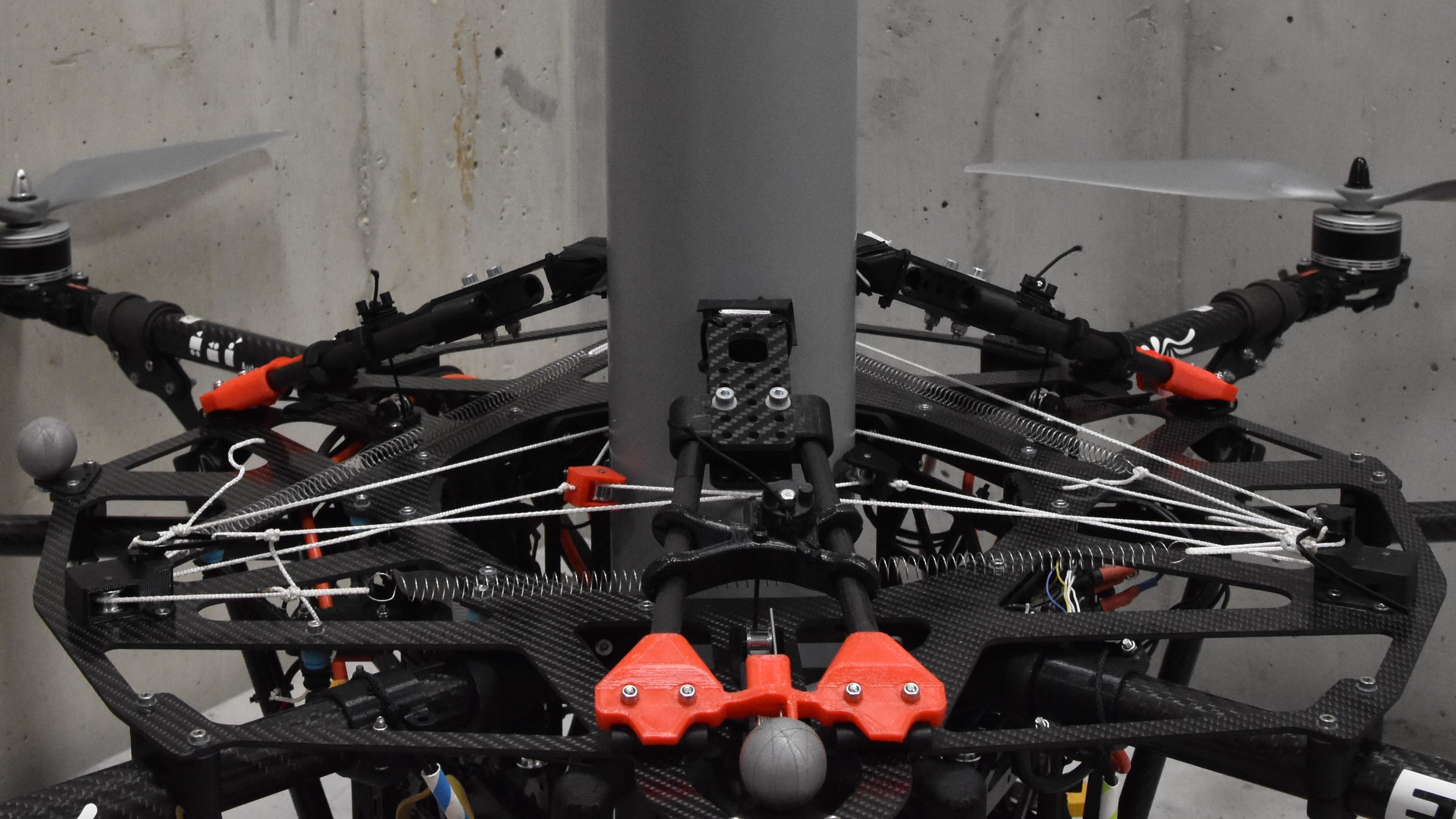}
        \else
        \includegraphics[width=\linewidth]{images/submission/high_quality/4f_lift_3.jpg} 
        \fi
        \caption{}
        \label{fig:gripping_full_f}
    \end{subfigure}
    \caption{\review{Full gripping procedure: Once the pole is located within the vertical clearance of  \Geranos{} (a), the gripping procedure begins. The centering mechanism aligns the pole (or rather the UAV, when airborne), such that the central axis of the pole is as close as possible to the  the $z$-axis of $\F_b$ (b). Once the centering procedure has finished (c), the self locking lifting mechanism comes in to play (d). By folding down all three hinged triangles (e), they will come in contact with the pole (f), effectively clamping it in place. When releasing the pole, this operation is done in reverse. This, however, only works if the pole is placed on the ground, since the self-locking mechanism can only be released if the weight of the pole is counteracted by an external force, e.g., the ground reaction.}}
    \label{fig:gripping_full}
\end{figure*}

In \review{\cref{fig:gripping_full} a-c}, the centering procedure is shown in three stages.
This mechanism is integrated twice on the UAV, once at the top, and once at the bottom, \qty{30}{\centi\meter} apart, \review{oriented in the same way,} such that the $z$-axis of $\F_b$ passes through the center of both mechanisms. Two centering mechanisms prevent the pole from translational movement in the $xy$-plane and rotation around any axis other than the $z$-axis of the body frame. 
\review{Due to possible manufacturing mismatches between the two centering systems, they would require two independent actuators to make sure that both are tight around the transported pole. However, to keep the system simple and lightweight, we utilize only one motor, mounted on a slightly compliant vertical element. This solution allows the motor itself to translate vertically of a few millimeters whenever needed, ensuring the cables of both the centering systems to be tight at the same time.}
%
%
The maximum radial misalignment of \Geranos{} relative to the pole is the difference between the radius of the pole $\poleradius$ and the radius of the incircle of the equilateral triangle $\inradius$. We chose $\inradius$ to be \qty{12.5}{\cm} and $\poleradius$ to be between \qty{5}{\cm} and \qty{7.5}{\cm}, resulting in a radial tolerance, $t = R - r_{max}$, of \qty{5}{\cm}. Those values were chosen to keep \Geranos{} on a relatively small scale and allow for small misalignments.
\subsubsection{Lifting}
To prevent the pole from moving along the $z$-axis of $\F_b$, we designed a friction-based self-locking mechanism. In \review{\cref{fig:gripping_full} d-f}, the mechanism is shown. 
Three hinged triangles are arranged at the vertices of an equilateral triangle. The angle $\foldangle \in \mathbb{R}$ between the $xy$-plane of $\F_b$ and the folding triangle, indicated in the red box in \cref{fig:geranos_sketch}, is initially at its maximum, ensuring that folding triangles do not obstruct the vertical clearance for the pole. \review{The folding triangles are actuated by elastic cables which pull the triangles down, meaning $\foldangle$ decreases, until the tips, equipped with a high friction rubber, of all three folding triangles are in contact with the pole. The three elastic cables are guided by pulleys to the motor. Elastic cables are used to ensure that all three folding triangles get in contact with the pole. Torsional springs push the triangles back up, once the payload is released.}
The relevant forces in one of the three folding triangles are also depicted in the red box in \cref{fig:geranos_sketch}.
The friction force $\f_f \in \mathbb{R}^{3}$, which acts in the positive $z$ direction of $\F_b$, counteracts the weight $\f_g \in \mathbb{R}^{3}$ of the pole. As there are three folding triangles, and $\f_g$ is the total weight of the pole, it holds: $\sum_{i=1}^3 \f_{f,i} = -\f_g$, where the subscript $i$ refers to the $i$-th folding triangle. Due to the inclination angle $\foldangle$, there is also a normal force $\f_n \in \mathbb{R}^{3}$ acting radially on the pole. If we consider all three folding triangles, the normal forces will cancel each other: $\sum_{i=1}^3 \f_{n,i} = \mathbf{0}$. 
\review{This holds if each folding triangle has the same inclination angle $\foldangle$. This is guaranteed by the centering mechanism assuming it perfectly centers the pole which must be cylindrical. 
In reality this cannot be achieved flawlessly. If one angle $\foldangle$ of one folding  triangle is different from the others, the lifting mechanism effectively exerts a torque on the pole, because the normal forces are no longer coplanar and do not have the same magnitude.
Thanks to the powerful actuation of the centering mechanism and its precise calibration, the difference between the three instances of $\foldangle$ and the resulting torques were negligible.
}\strike{To further discuss the relation of these forces, we will only consider the forces in one folding triangle. There are}
\review{To demonstrate the self-locking capabilities of our mechanism, consider the} two relations between the normal force $\f_n$ and the friction force $\f_f$. First, the sum of $\f_n$ and $\f_f$ has to be a force which is parallel to the folding triangle. This results in the relation: $|\f_f| = \tan(\foldangle) |\f_n|.$
Second, the no slip condition between the pole and the tip of the folding triangle has to be fulfilled, given the static friction coefficient $\frictioncoeff \in \mathbb{R}$: $|\f_f| \le \frictioncoeff |\f_n|.$
Combining these relations gives a condition between the static friction coefficient $\frictioncoeff$ and the angle $\foldangle$:
\begin{equation}
    \frictioncoeff \geq \tan \foldangle.
    \label{eq.:friction}
\end{equation}
Note that no forces are involved in this relation, which is why we call this mechanism self-locking. Thus, even for arbitrarily large $|\f_g|$, as long as the structural integrity of the UAV is ensured and the relation in \eqref{eq.:friction} holds, the load can be lifted. 
To ensure that this condition is met, according to the pole diameter, we can manually adjust the length of the folding triangle which results in a different value for $\foldangle$. 
%
\review{
\subsubsection{Gripper Actuation}
The actuation of the full gripping mechanism was achieved by two servo motors with adjustable output torque. The actuator for the centering mechanism is programmed to exert a torque throughout the full duration of the pole's attachment to the UAV. The actuation of the lifting mechanism is only required when grabbing and releasing the payload. A control input (autonomous or human-based) initiates the grasping or releasing procedure. 
When grabbing, first the centering mechanism is actuated, followed by the lifting mechanism, as in~\cref{fig:gripping_full}. When releasing, the order is reversed. Essentially, the lifting mechanism is only active, when the centering mechanism is engaged.   
}



\subsection{\review{Propeller} Actuation Setup}
To further mitigate the impact of the inertia of the pole, we use a rotor configuration that enables the UAV to keep its angular acceleration small by not tilting around $x$ and $y$ axes of $\F_b$.
This cannot be obtained with standard collinear multi-rotor vehicles because of their underactuation.
We therefore designed a different propeller setup.

\Geranos{} has four \qty{20}{\inchQ} main propellers, arranged in a traditional collinear quadcopter setup and four \qty{5}{\inchQ} auxiliary propellers, tilted outwards as shown in the green box in \cref{fig:geranos_sketch}, to enable lateral motion without tilting. All auxiliary propellers lie on the $xy$ plane of $\F_b$ and are connected to the body by tubular carbon-fibre rods. The airflow is directed radially away from the body to mitigate unwanted aerodynamic effects. The main rotors are linked to the hexagonal prism by a triangular truss structure, consisting of carbon rods and panels, to reduce internal stresses. 

\section{System Modeling}
\label{sec:SystemModeling}

To take advantage of model-based control synthesis, we modeled the UAV \Geranos{} similar to~\cite{FullPoseOmnidirectionality}.
We make the following assumptions to simplify the modeling of the system:
\begin{enumerate*}[label=\roman*)]
    \item The body of the UAV is rigid;
    \item The thrust and the torque produced by each propeller are proportional to the propeller's squared angular velocity;
    \item No aerodynamic interference between propellers is considered as the slipstreams in the configuration of the rotors do not cross;
    \item The \review{aerodynamic} interaction with the ground and other surfaces is neglected;
    \item The aerodynamic drag of the main body is negligible due to its small linear and angular velocity.
\end{enumerate*}
These simplifications lead to only a slight model uncertainty that the controller can compensate.

The frame with respect to which a vector or matrix is defined, is indicated using the conventional notation of a trailing subscript. Thus, $\star_b$ represents a generic vector $\star$ expressed in frame $\F_b$. 



We use the Newton-Euler formalism to derive \Geranos{}' dynamics with respect to $\F_b$:

\begin{align}
    \begin{split}
        \begin{bmatrix}
            {m}\mathbb{I}_3 & 0 \\
            0 & \mathbf{J}_b \\
        \end{bmatrix}
        \begin{bmatrix}
            \dot{\mathbf{v}}_b \\
            \dot{\boldsymbol{\omega}}_b
        \end{bmatrix}
        +
        \begin{bmatrix}
            \boldsymbol{\omega}_b \times m \mathbf{v}_b\\
            \boldsymbol{\omega}_b \times \mathbf{J}_b \boldsymbol{\omega}_b
        \end{bmatrix}
        = &\\
        \begin{bmatrix}
            \mathbf{f}_{b} \\
            \boldsymbol{\tau}_{b}
        \end{bmatrix}
        - &
        \begin{bmatrix}
            m \mathbf{R}_{wb}^\top \mathbf{g}_w \\
            \mathbf{x}_{com} \times \mathbf{f}_{b}
        \end{bmatrix}
    \end{split},
    \label{eq:system_model}
\end{align}
where, \review{$\mathbb{I}_n \in \mathbb{R}^{n\times n}$ is the identity matrix of dimension $n$}. Furthermore, $m  \in \mathbb{R}$ is the overall mass of the system, $\boldsymbol{J}_{b}  \in \mathbb{R}^{3\times 3}$ is its inertia matrix and $\mathbf{p}_w\in \mathbb{R}^{3}$, $\mathbf{v}_b\in \mathbb{R}^{3}$ and $\boldsymbol{\omega}_b\in \mathbb{R}^{3}$ are its position, linear velocity and angular velocity. Additionally, $\mathbf{f}_b\in \mathbb{R}^{3}$ and $\boldsymbol{\tau}_b\in \mathbb{R}^{3}$ are the total actuator force and torque vectors acting on the geometric center of the system. $\mathbf{R}_{wb}\in SO(3)$ is the rotation matrix from $\F_b$ to $\F_w$ and $\mathbf{g}_w \review{\in \mathbb{R}^{3}}$ is the gravity vector along the $z$-axis of the world frame. 
As in \cite{FullPoseOmnidirectionality}, the term $\mathbf{x}_{com} \times \mathbf{f}_{b}$ is subtracted to account for an offset $\mathbf{x}_{com}\in \mathbb{R}^3$ between the geometric center of \Geranos{} and its \com{}. \review{The inertia matrix $\boldsymbol{J}_{b}$ and the \com{}-offset $\mathbf{x}_{com}$ were computed with a computer-aided design (CAD) model of \Geranos{}.}

As a result of previous assumptions, the force and torque generated by the $i$-th propeller acting on the geometric center of the UAV, $\mathbf{f}_{b,i}\in\mathbb{R}^3$ and $\boldsymbol{\tau}_{b,i}\in\mathbb{R}^3$, are approximated by: 
\begin{equation}
    \begin{split}
        \mathbf{f}_{b,i} &= \underbrace{c_{fi}  \boldsymbol{\xi}_{b,i}}_{\boldsymbol{\beta}_{b,i}}  \, w_i = \boldsymbol{\beta}_{b,i} \, w_i 
        \\
        \boldsymbol{\tau}_{b,i} &= \underbrace{\left(c_{fi}  \textbf{r}_{b,i} \times \boldsymbol{\xi}_{b,i} + c_{Mi} \boldsymbol{\xi}_{b,i}\right)}_{\boldsymbol{\gamma}_{b,i}}w_i = \boldsymbol{\gamma}_{b,i} \, w_i.
    \end{split}
    \label{eq:propeller_torque_force}
\end{equation}
Here $c_{fi} \in \mathbb{R}$ is the propeller specific constant which relates the squared angular velocity $w_i \in \mathbb{R}$ of the propeller $i$ to the force generated by the propeller. 
Likewise, $c_{Mi} \in \mathbb{R}$ is the constant that relates the squared angular velocity to the drag moment produced by the propeller. \review{$c_{fi}$ and $c_{Mi}$ were evaluated with a force and torque sensor for each propeller type on a testing rig.}
$\boldsymbol{\xi}_{b,i}\in\mathbb{R}^{3}$ is the unit vector perpendicular to the propeller plane, and $\textbf{r}_{b,i}\in\mathbb{R}^{3}$ is the position of the propeller, \review{both expressed in $\F_b$}.
$\boldsymbol{\beta}_{b\review{,i}}\in\mathbb{R}^{3}$ and $\boldsymbol{\gamma}_{b\review{,i}}\in\mathbb{R}^{3}$ are therefore constant vectors.
The stacked vector of force $\mathbf{f}_b$ and torque $\boldsymbol{\tau}_b$ in \eqref{eq:system_model} produced by all propellers is called wrench and denoted as $\mathbf{u}\in\mathbb{R}^6$. The wrench can be expressed by the following:

\begin{align}
    \begin{split}
        \mathbf{u}_b = 
        \begin{bmatrix}
            \mathbf{f}_b \\
            \boldsymbol{\tau}_b
        \end{bmatrix}
        = \sum_{i=1}^{8}
        \begin{bmatrix}
            \mathbf{f}_{b,i} \\
            \boldsymbol{\tau}_{b,i}
        \end{bmatrix}
        = \mathbf{A}_b \mathbf{w}
    \end{split}
    \label{eq:wrench},
\end{align}
where $\mathbf{A}_b\in\mathbb{R}^{6\times 8}$ is denoted as \textit{allocation matrix} and $\mathbf{w} = [w_1 \; \dots \; w_8]^\top \in \mathbb{R}^8$ is the vector of the squared angular velocities of all eight propellers.
\strike{where $\mathbf{A}_b\in\mathbb{R}^{6\times 8}$ is denoted as \textit{allocation matrix} and the index $i$ indexes all eight propellers of \Geranos{}. $\mathbf{w} = [w_1 \; \dots \; w_8]^\top \in \mathbb{R}^8$ in equation \eqref{eq:wrench} is the vector of the squared angular velocities of all eight propellers.}

As seen in \eqref{eq:wrench}, $\mathbf{A}_b$ maps the squared angular velocities to the force and torque on the \com{}. Thanks to \review{\Geranos{}'} propeller setup, \review{$\mathbf{A}_b$} has full rank making the system fully actuated~\cite{2021-HamUsaSabStaTogFra}.
\review{However, the system cannot hover in any orientation (it is not omnidirectional) because the auxiliary propellers are not strong enough to fully counterbalance \Geranos{}' weight in non-flat orientations}\footnote{We denote a flat orientation as an orientation of the UAV, where the z-axes of $\F_b$ and $\F_w$ are aligned.}. Therefore, as long as \review{\Geranos{}} is close to a flat orientation, it can independently control the orientation and translation dynamics in all directions, enabling it to fly horizontally without tilting its body. 

The propeller setup also results in the vehicle being over-actuated since $\mathbf{A}_b$ has a 2-dimensional nullspace. 
\review{This enables the vehicle to maintain full actuation while hovering, even when there are limitations such as the maximum angular velocities or preferred spinning directions of the motors. Additionally, this setup allows for optimal energy consumption.}
\strike{This enables the full actuation to be maintained \review{around hovering} despite restrictions like the motors' maximum angular velocities or preferred spinning directions and allows for the energy consumption to be optimized.}
\section{Control}

\label{sec:control}


\tikzstyle{block} = [draw, fill=gray!20, rectangle, 
    minimum height=3em, minimum width=6em]
\tikzstyle{sum} = [draw, fill=gray!20, circle, node distance=1cm]
\tikzstyle{input} = [coordinate]
\tikzstyle{output} = [coordinate]
\tikzstyle{pinstyle} = [pin edge={to-,thin,black}]

\begin{figure}[t]
    \centering
    \ifarxiv
    \includegraphics[width=\linewidth]{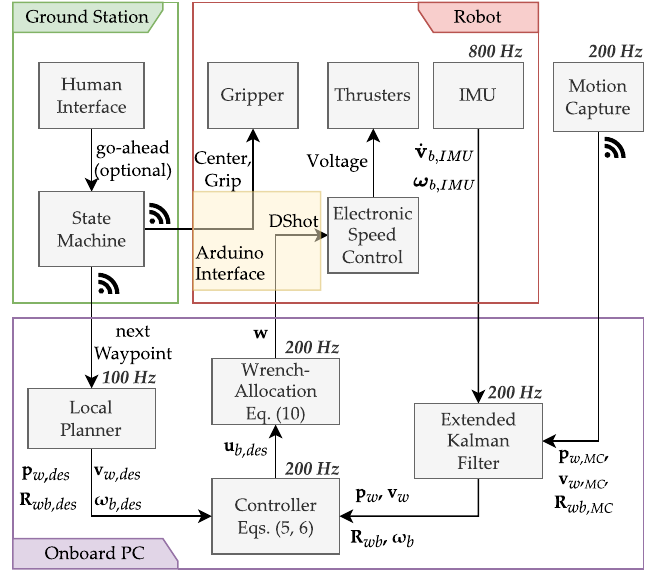} 
    \else 
    \includegraphics[width=\linewidth]{images/submission/eps/5_Geranos_Controller_Figure_final.eps} 
    \fi
    \caption{\review{System Architecture of \Geranos{}. The System comprises of three main categories: The commands sent from the ground station, a control loop running the onboard computer, and the sensors and actuators mounted on the robot. To estimate the robot's state, sensor measurements from an onboard IMU and an external motion capture system are fused with an Extended Kalman Filter. In case a node runs periodically, its frequency is specified above it.}}
    \label{fig:controller_structure}
\end{figure}

\review{The \Geranos{} controller is adapted from~\cite{FullPoseOmnidirectionality}. An overview of the high- and low-level system architecture can be seen in~\cref{fig:controller_structure}. The controller loop takes the desired pose and twist (subscript $des$) as input, which are set by a local planner sampling from a trajectory at \qty{100}{\Hz}. This trajectory is created by interpolating waypoints, set by the state machine, with a $9^{th}$ order polynomial trajectory. The state machine determines the order of high-level tasks (e.g., lift pole, then place pole) and sets the waypoints required to fulfill these tasks accordingly.}


\review{The controller computes the wrench target $\mathbf{u}_{b,\review{des}} = [\mathbf{f}_{b,des}^\top, \boldsymbol{\tau}_{b,des}^\top]^\top \in \mathbb{R}^6$ based on the state error (pose and twist).
$\mathbf{u}_{b,\review{des}}$ is then processed by the allocation method to get the propeller's squared angular velocities $\mathbf{w}$. These velocities are transmitted to the Electronic Speed Controls using the Digital Shot (DShot) communication protocol.}

\subsection{Controller}
\label{sec:controller}
\review{
The controller calculates the wrench target $\mathbf{u}_{\review{b,des}}$ according to the deviation from the desired pose and twist. We define $\textbf{e}_\star{}\in\mathbb{R}^3$ as the errors in position ($\star{} = p$), velocity ($\star{} = v$), attitude ($\star{} = R$) and angular velocity ($\star{} = \omega$).}
\begin{equation}
    \begin{aligned}
        \textbf{e}_p & = \mathbf{p}_w - \mathbf{p}_{w,des} \\
        \textbf{e}_v & = \mathbf{v}_w -\mathbf{v}_{w,des} \\
        \textbf{e}_R & = \frac{1}{2}\left(\mathbf{R}_{wb,des}^\top \mathbf{R}_{wb} - \mathbf{R}_{wb}^\top \mathbf{R}_{wb,des}\right)^\vee \\
        \textbf{e}_\omega & = \boldsymbol{\omega}_b - \mathbf{R}_{wb}^\top \mathbf{R}_{wb,des} \boldsymbol{\omega}_{b,des}, \\
    \end{aligned}
    \label{eq:errors}
\end{equation}
whereby $(...)^\vee$ is the vee-map, which transforms a skew-symmetric matrix into a 3-dimensional vector. The inverse operation $(...)^\wedge$ is the hat-map. 

From the errors in \eqref{eq:errors}, the wrench target is computed by a proportional-derivative action plus feedforward and dynamic cancellation: 

\begin{equation}
    \begin{aligned}
    \f_{b,des} &= m (\mathbf{R}_{wb}^\top(-k_p \cdot \mathbf{e}_p - k_v \cdot \mathbf{e}_v - k_i \cdot \mathbf{e}_i \\
    & \qquad+ \mathbf{\dot{v}}_{w,des} + \mathbf{g}_w)+(\boldsymbol{\omega}_b^\wedge \mathbf{R}_{wb}^\top \mathbf{v}_w)) \\
    \boldsymbol{\tau}_{b, des} &= \mathbf{J_b}(-k_R \mathbf{e}_R-k_\omega \mathbf{e}_\omega)+(\boldsymbol{\omega}_b^\wedge \mathbf{J_b} \boldsymbol{\omega}_b)\\
    & \qquad+(\mathbf{x}_{com}^\wedge \review{\mathbf{f}_{b,des}}).   
    \end{aligned}
    \label{eq:controller}
\end{equation}

Here, \review{$k_{p}$, $k_{v}$, $k_{i}$, $k_R$ \text{ and } $k_{\omega}$ $\in\mathbb{R}$} are constant PID-gains. 
To account for a static position errors, we add an integral error $\mathbf{e}_i\in\mathbb{R}^3$ for the calculation of the desired force $\f_{b,des}\in\mathbb{R}^3$ in \eqref{eq:controller}. \review{The integral position error $\mathbf{e}_i$ is numerically integrated from the position error $\mathbf{e}_p$ and saturated at a lower and upper bound of $\pm \SI{1}{\meter\second}$.}
\review{In our setting, the desired attitude is a constant flat orientation, aiding the assembly task. Therefore,  the auxiliary propellers' sizing determines the largest side-force that \Geranos{} may withstand.
However, the use of an integral term could cause problems during the grasping and releasing phases of the transported poles. For this reason we pause the accumulation of the integral term during this time, allowing the platform to be compliant with respect to the poles.}

\subsection{\review{Wrench-}Allocation}
\label{sec:allocation}

One of the main challenges that arose when designing a multi-rotor vehicle like \review{\Geranos{}}, is allocating the desired wrench computed by the controller in \eqref{eq:controller} into the actuators of the UAV. 
Although the system is fully-actuated, the allocation problem gets non trivial when considering the actuation limits. Each propeller $i$'s angular velocity is constrained by a minimal and maximal value, \review{$w_{i,min}, w_{i,max} \in \mathbb{R}_{\geq0}$, different for main and auxiliary propellers}. Therefore, the following element-wise relation must always hold:
\review{
\begin{align}
    \begin{split}
    \mathbf{w}& \geq \mathbf{w}_{min} = [w_{min,1}, \dots, w_{min,8}]^\top \\
    \mathbf{w}& \leq \mathbf{w}_{max}= [w_{max,1}, \dots, w_{max,8}]^\top.
    \end{split}
    \label{eqn:input_limits}
\end{align}
}
%
As the allocation matrix has a 2-dimensional null-space, there is an infinite amount of viable mappings. 
\review{We decided to solve this problem using optimal control. 
In particular, the angular velocities are found as the solution of the minimization of the weighted $l_2$-norm of the vector $\mathbf{w}$ (energy efficiency) under the constraint $\mathbf{A}_b\cdot\mathbf{w} = \mathbf{u}_{\review{b,des}}$ (generation of the wrench target), as well as the input limits in \eqref{eqn:input_limits}.
We solve this optimization problem with quadratic programming (QP), as in~\cite{2020a-NavSabTogPucFra}.}

We decided to add \review{a vector of} slack variables $\pmb{\delta}\in\mathbb{R}^6$, i.e., we augmented the QP problem, in order to guarantee that the \review{problem stays feasible}. This is required \review{for safety reasons, to let} the solver to handle any wrench target.
The equality constraint thus becomes:

\begin{equation}
        \textbf{A}\review{_b}\cdot\mathbf{w} + \pmb{\delta} = 
        \begin{bmatrix}
            \textbf{A}\review{_b} & \mathbb{I}\review{_6} \\
        \end{bmatrix}\cdot 
        \underbrace{
        \begin{bmatrix}
            \mathbf{w} \\
            \pmb{\delta}
        \end{bmatrix}}_{\textbf{y}} = \begin{bmatrix}
            \textbf{A}\review{_b} & \mathbb{I}\review{_6} \\
        \end{bmatrix}\cdot \textbf{y} = \textbf{u}_{\review{b,}des}.
\end{equation}
The minimization variable is the vector $\mathbf{y}\in\mathbb{R}^{14}$ which is also constrained by a lower bound \review{$\mathbf{y}_{lb} \in \mathbb{R}^{14}$} and \review{an} upper bound \review{$\mathbf{y}_{ub} \in \mathbb{R}^{14}$}:
\begin{equation}
\begin{aligned}[t]
    \mathbf{y}_{lb} &= 
    \begin{bmatrix}
        \mathbf{w}_{min} \\ -\mathbf{1}_{6} \cdot \zeta
    \end{bmatrix}\\
\end{aligned}
\qquad 
\begin{aligned}[t]
    \mathbf{y}_{ub} &= 
    \begin{bmatrix}
        \mathbf{w}_{max} \\ \mathbf{1}_{6} \cdot \zeta
    \end{bmatrix},
\end{aligned}
\end{equation}
where $\mathbf{1}_{n}\in\mathbb{R}^n$ denotes the vector filled with ones and $\zeta \approx 10^6$ is a large number bounding the slack variables. 
The QP-problem can be formulated as follows.

\begin{equation}
    \begin{aligned}
    \mathbf{y}^\star = \argmin_{\mathbf{y}} \, \mathbf{y}^\top \, \mathbf{H}\, \mathbf{y} \\
    \text{Subject to: }\hspace{1mm} \begin{bmatrix}
            \mathbf{A}\review{_b} & \mathbb{I}\review{_6} \\
        \end{bmatrix}\cdot\mathbf{y} & = \mathbf{u}_{\review{b,}des} \\
        \mathbf{y} & \geq \mathbf{y}_{lb} \\
        \mathbf{y} & \leq \mathbf{y}_{ub} \\
\end{aligned},
\label{eq:qp}
\end{equation}
where $\mathbf{H}\review{ = \text{diag}([\mathbf{h}_{\mathbf{w}, main}^\top, \mathbf{h}_{\mathbf{w}, aux}^\top, \mathbf{h}_{\boldsymbol{\delta}}^\top])}\in\mathbb{R}^{14\times 14}$ is a diagonal matrix weighting each value of $\mathbf{y}$.
\review{Each of the vectors 
$\mathbf{h}_{\mathbf{w}, main} = h_{\mathbf{w}, main} \cdot \mathbf{1}_4 \in \mathbb{R}^4$, $\mathbf{h}_{\mathbf{w}, aux} = h_{\mathbf{w}, aux} \cdot \mathbf{1}_4 \in \mathbb{R}^4$, and
$\mathbf{h}_{\boldsymbol{\delta}} = h_{\boldsymbol{\delta}} \cdot \mathbf{1}_6 \in \mathbb{R}^6$ weighs the corresponding elements of vector $\mathbf{y}$ in the QP-Problem. As such, $h_{\mathbf{w}, main}$ is the weight corresponding to the angular velocities of the main propellers, $h_{\mathbf{w}, aux}$ is the weight corresponding to the angular velocities of the auxiliary propellers and $h_{\boldsymbol{\delta}}$ is the weight of the slack variables $\boldsymbol{\delta}$. We set $h_{\mathbf{w}, main} = 1$ as a reference. We weight the squared angular velocities of the auxiliary propellers $4$ times higher than the squared angular velocities of the main propellers $h_{\mathbf{w}, aux} = 4 \cdot h_{\mathbf{w}, main}$ in order to avoid reaching the actuators' saturation.}

\review{To determine the weight $h_{\boldsymbol{\delta}}$ associated to the slack variables $\boldsymbol{\delta}$, we first acknowledge the difference in units between the slack variables $\pmb{\delta}$ (in $[\qty{}{\newton}] \, \text{or} \,[\qty{}{\newton\meter}]$) and the squared angular velocities $\mathbf{w}$ (in $[\qty{}{\radian\squared\per\second\squared}]$).
On average, the squared angular velocities are approximately $10^6$ larger than the slack variables. Additionally, we want to prioritize $\pmb{\delta}$ by assigning it a significantly higher weight after adjusting for scale, which we have empirically determined to be a factor of $10^7$.
This decision is made to minimize wrench errors that could negatively impact tracking performance and potentially lead to costly system failures. Ultimately, the slack variables are weighted with a total factor of $h_{\boldsymbol{\delta}} = 10^{13} \cdot h_{\mathbf{w}, main}$}.

\subsection{Inclusion of Pole Dynamics}
\label{sec:pole_dynamics}

A major advantage of grabbing a load rigidly is that the inertia and mass of the load can be included in the model of the system and therefore do not have to be treated as a disturbance.
We do this by changing the mass and inertia of the system as soon as the rigid connection to the payload is established. For this, we assume that the physical properties of the payload (mass and inertia) are known and the robot grabs the payload at approximately the same position every time.
\review{In practice, to estimate the lifted poles' inertia, we modeled the poles as cylindrical tubes.}

To calculate the combined inertia we first assume that the principal axes of the load are aligned with the body frame and utilize the parallel axis theorem to determine the inertia of the load with respect to the body frame $\mathbf{J}_{b,load}$ $\review{\in \mathbb{R}^{3\times 3}}$:

\begin{equation}
    \mathbf{J}_{b,load} = \mathbf{J}_{com,load} - m_{load} \left(\mathbf{d}^\wedge\right)^2.
\end{equation}
Here $\mathbf{J}_{com,load}$ $\review{\in \mathbb{R}^{3\times 3}}$ is the load's moment of inertia relative to its \com{}. $\mathbf{d}\in\mathbb{R}^3$ is the vector from the center of the body frame to the load's \com{}. Thanks to the ability of \Geranos{} to grab a load close to its \com{}, $\mathbf{d}$ is minimized. We recall that $\mathbf{J}_b$ and $\mathbf{J}_{b,load}$ are both relative to the body frame and therefore cumulative. Hence, we can simply add $\mathbf{J}_{b,load}$ to the current inertia.

\strike{The physical characteristics are changed in a linear fashion so as to prevent commanding a non-continuous input.}
Although this method of updating the physical characteristics causes a \review{fast} change in the dynamics of the system, we did not encounter any problems using this approach, presumably because the robot is in a quasi-static condition \review{during the grasping phase}.


\section{Experiments and Results}

\strike{We validated the concept presented in the previous sections on a real prototype. In the following, we present some implementation details of the prototype as well as the experiments we used to evaluate the concept.}

\subsection{Hardware Setup}
\label{sec:hardware}

\Geranos{} is equipped with four T-Motor MN505-S \qty{320}{\KV} motors with \qtyproduct{20 x 10}{\inchQ} propellers for the main thrust, as well as four T-Motor F80 Pro \qty{1900}{\KV} motors with \qtyproduct{5.1 x 4.6}{\inchQ} propellers that are used for the side thrust.
The motors are driven with two F55A PRO II 6S 4IN1 Electronic Speed Controllers (ESCs). The gripping mechanism uses two Dynamixel XL430-W250 motors with the U2D2 controller board. Furthermore, the vehicle is equipped with a FrSky \qty{2.4}{\giga\hertz} ACCST R-XSR RC-Receiver and a VN-100 \review{inertial measurement unit (IMU)} from VectorN\review{av}. In its current stage of development, an additional pose measurement is provided by a VICON motion capture system (MOCAP). To power the complete system two 6S1P \qty{22.2}{\V} \qty{6200}{\mAh} LiPo batteries are used, combined with a custom-made power distribution board with integrated voltage regulators for the Dynamixel motors and the on-board computer, for which an Udoo Bolt V8 is used. 

With the T-Motor MN505-S \qty{320}{\KV} as main thrust motors, we are able to achieve a maximum thrust of \qty{4.53}{\kg} per motor. Therefore, the thrust to weight ratio is 1.61 as \Geranos{}' overall weight carrying a \qty{3}{\kg} payload is \qty{11.26}{\kg}.

\subsection{Software Architecture} 
We built our framework on top of the widely used Robot Operating System (ROS)\strike{\cite{ros}}. The software architecture is split in two main components which are also shown in~\cref{fig:controller_structure}:
\subsubsection{High-level}
To provide a good estimation of the system's full pose, we use a sensor fusion based on an extended Kalman filter. 
This sensor fusion combines the filtered IMU data as well as the external pose measurements from \review{the VICON MOCAP}, each delivering the state at \qty{200}{\Hz}. 
Trajectories are then sent over a Wi-Fi network to the on-board computer. As the system's dynamics change when there is a payload attached, we implemented a state machine for the different flight stages, which changes model parameters, e.g., the weight, inertia, \com{} position, as well as controller gains. 

\subsubsection{Low-level}
The desired angular velocities computed by the controller in \cref{sec:control} are sent over a serial link to the Arduino, which communicates with the ESCs over the DShot protocol. This protocol requests, similar to pulse width modulation (PWM), a percentage of the maximum available motor velocity.
\review{During testing, we noticed that as the battery voltage decreased, the flight stability also decreased. This was because the true angular velocity of the propeller was lower than the commanded velocity. To address this issue and minimize uncertainty in the model, we introduced a battery voltage compensation. This compensation involved using a mapping technique inspired by \cite{VoltageCompensation} to determine the appropriate DShot motor commands based on the desired velocity and current battery voltage. This mapping was created by fitting a 3rd degree polynomial to a wide range of recorded data on battery voltages, DShot values, and propeller angular velocities.}

\subsection{Gripper Performance}
\label{sec:gripperperformance}


We tested the rigidity of the connection between \Geranos{} and the \review{load during flight} by monitoring the relative transformation between both objects \review{using the VICON MOCAP system}. We conducted four different \review{experiments, during which} we applied different accelerations to the system. \review{Simultaneously, we measured the misalignment in position and rotation between the CoM of the UAV and the pole. For the rotational error, the yaw offset was neglected since the vertically-transported poles are symmetrical in relation to their yaw direction.}

\begin{figure}[t]
    \centering
    \ifarxiv
    \includegraphics[width=\linewidth]{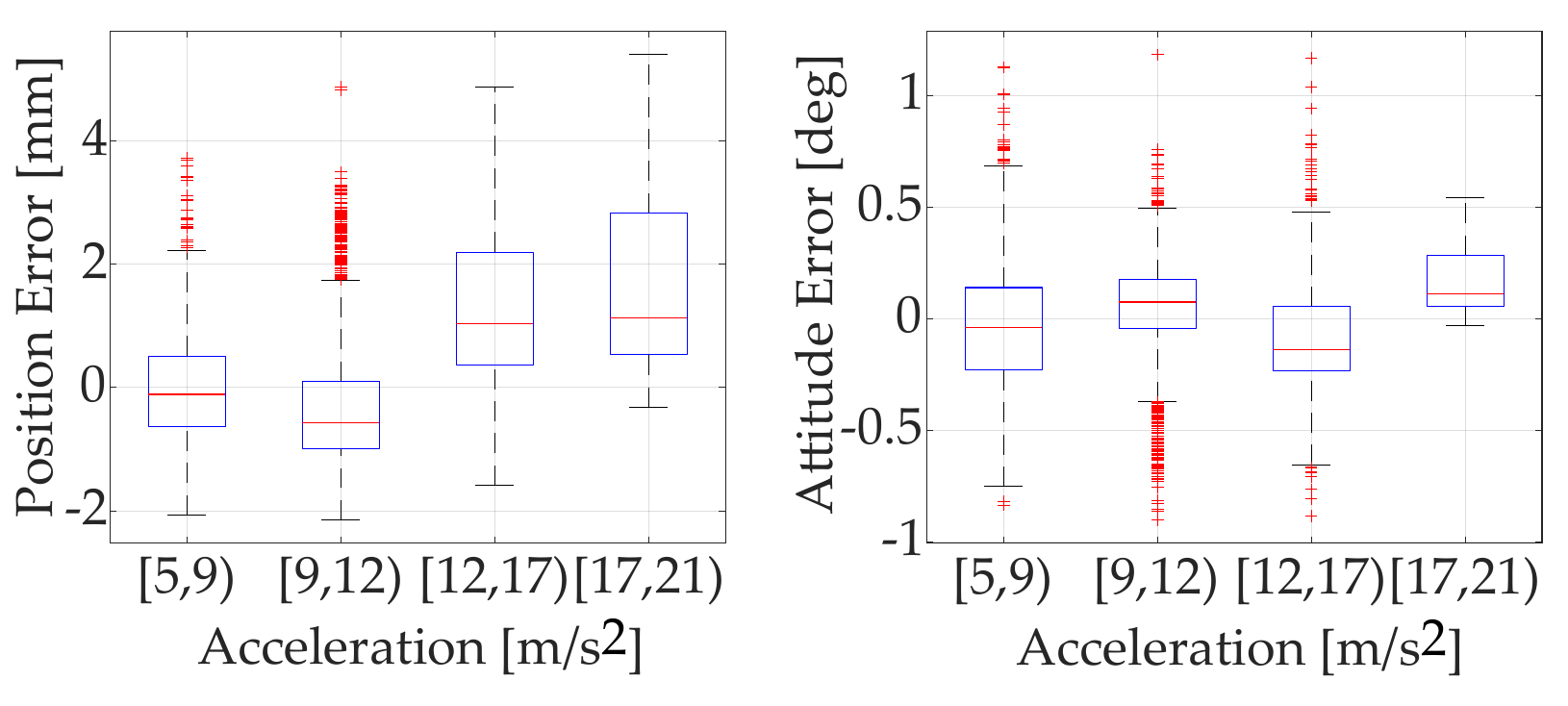}  
    \else
    \includegraphics[width=\linewidth]{images/submission/eps/6_boxplots_fig6.eps}
    \fi
    \caption{Position and attitude misalignment between \Geranos{} and the pole depending on the applied acceleration.}
    \label{fig:rigid_connection}
\end{figure}

\review{
According to the data presented in \cref{fig:rigid_connection}, even when the system experiences a high acceleration above \qty{12}{\meter\per\second\squared} or is vigorously shaken, the average position error remains within \qty{0.6}{\milli\meter} and the maximum attitude error is around \qty{1}{\degree}. These errors are considered insignificant, considering the size of the system. In our system demonstrations, we rarely encounter accelerations higher than \qty{12}{\meter\per\second\squared}, resulting in a maximum position error below \qty{4}{\milli\meter} and a maximum attitude error below \qty{1.2}{\degree}.
It should be noted that position errors exceeding \qty{5}{\milli\meter} only occur when the system experiences accelerations surpassing \qty{17}{\meter\per\second\squared}. Based on these findings, we can conclude that our gripper establishes a nearly rigid connection between Geranos and the pole. Furthermore, as the UAV only flies horizontally and the pole is held as close as possible to its center of mass (CoM), the torque applied to the pole is negligible, regardless of any rotational offset.
}

\subsection{Demonstration}
\label{sec:demo}

\begin{figure*}[t]
    \begin{minipage}[b]{\textwidth}
        \centering
        \ifarxiv
        \includegraphics[width=.32\linewidth]{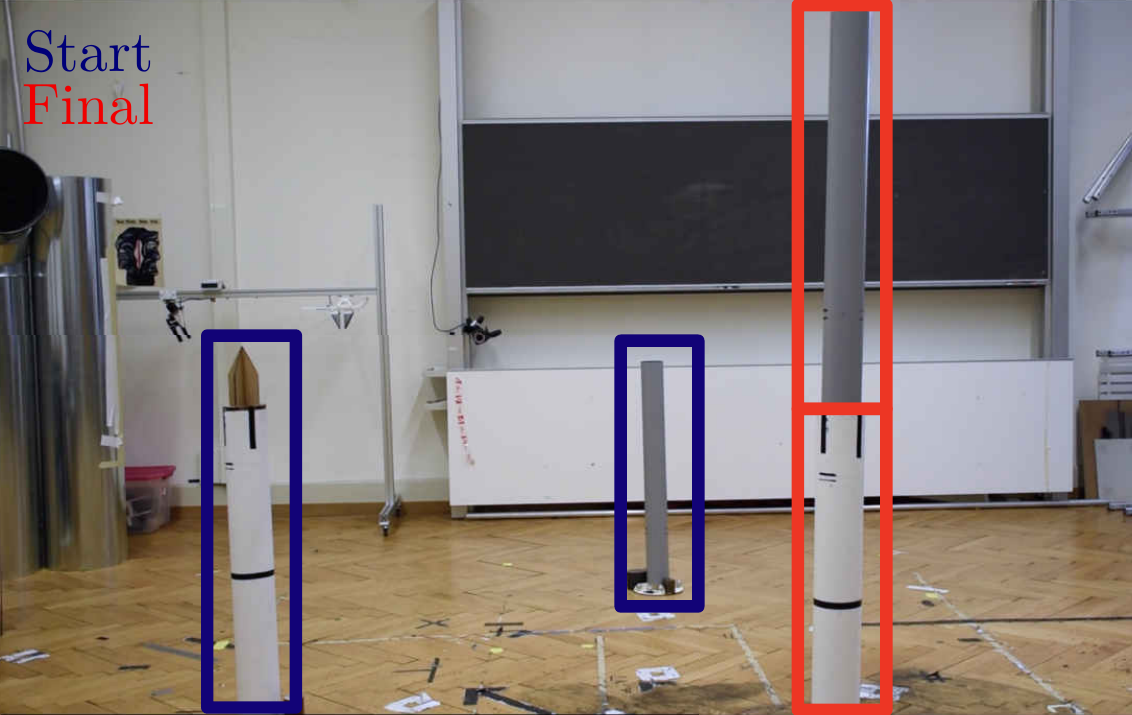}
        \includegraphics[width=.32\linewidth]{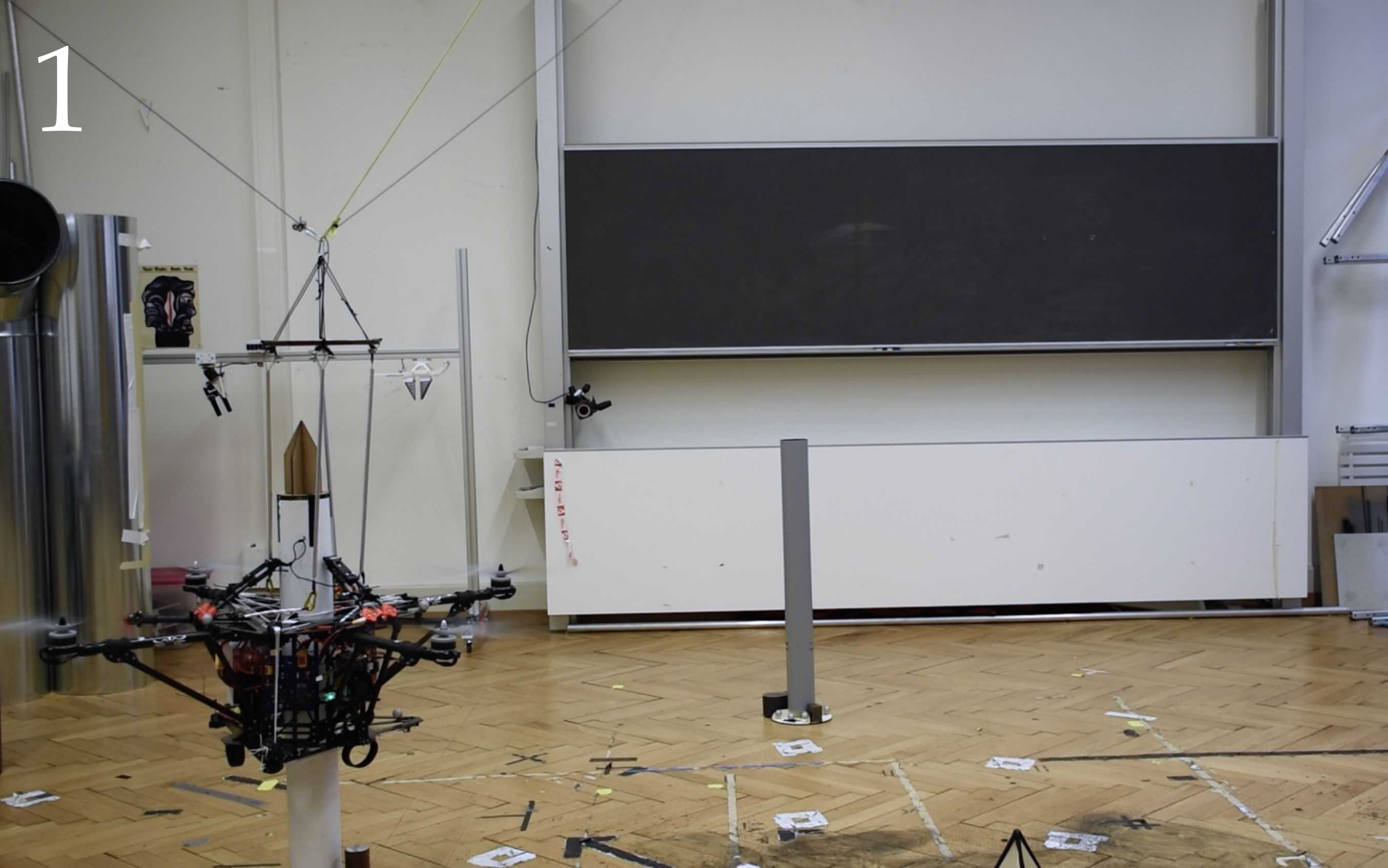}
        \includegraphics[width=.32\linewidth]{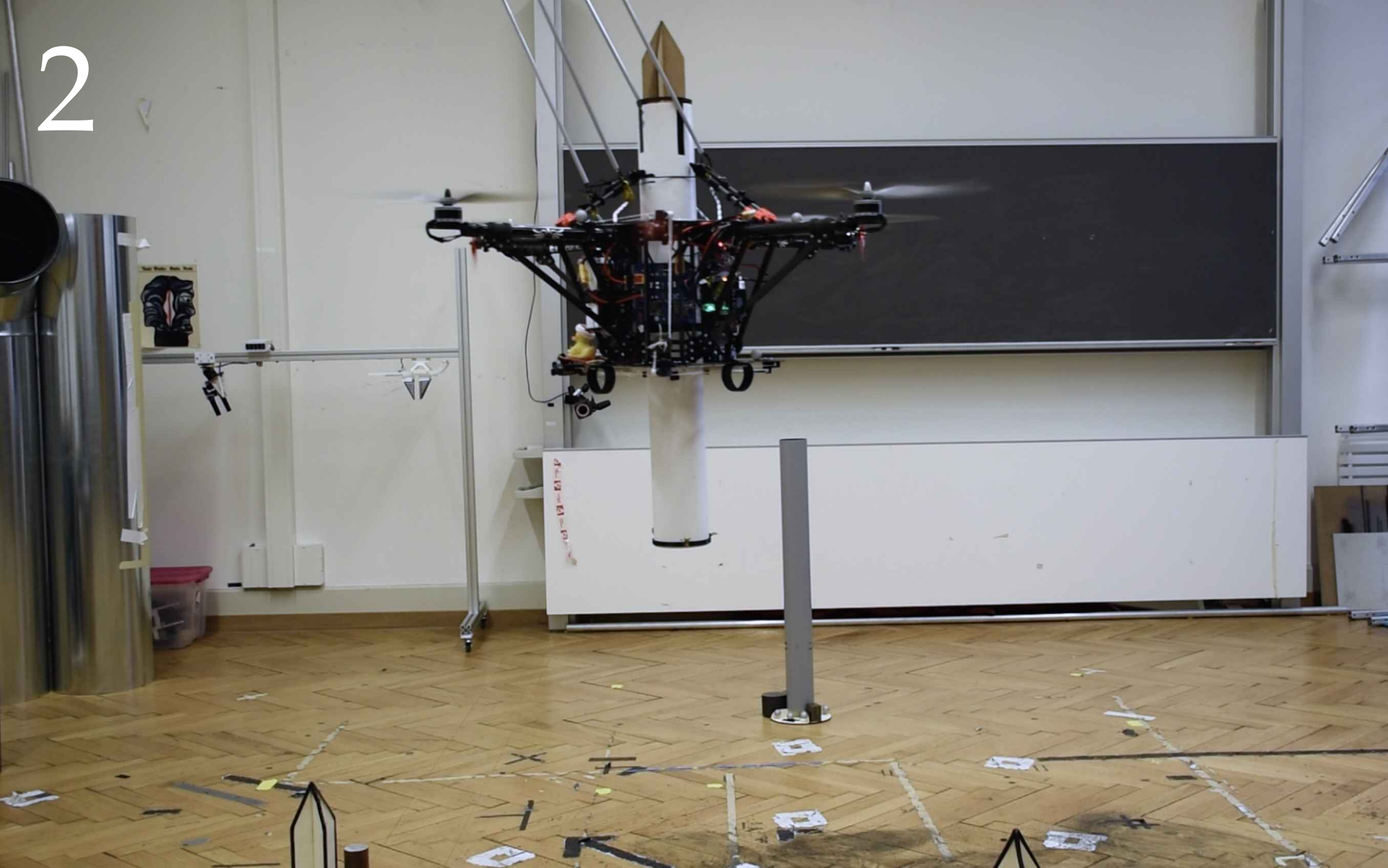}
        \else
        
        \includegraphics[width=.32\linewidth]{images/submission/eps/7a_image_poles.eps}
        \includegraphics[width=.32\linewidth]{images/submission/eps/7b_image1.eps}
        \includegraphics[width=.32\linewidth]{images/submission/eps/7c_image2.eps}
        \fi
    \end{minipage}
    
    \vspace{0.1cm}
    \begin{minipage}[b]{\textwidth}
        \centering
        \ifarxiv
        \includegraphics[width=.32\linewidth]{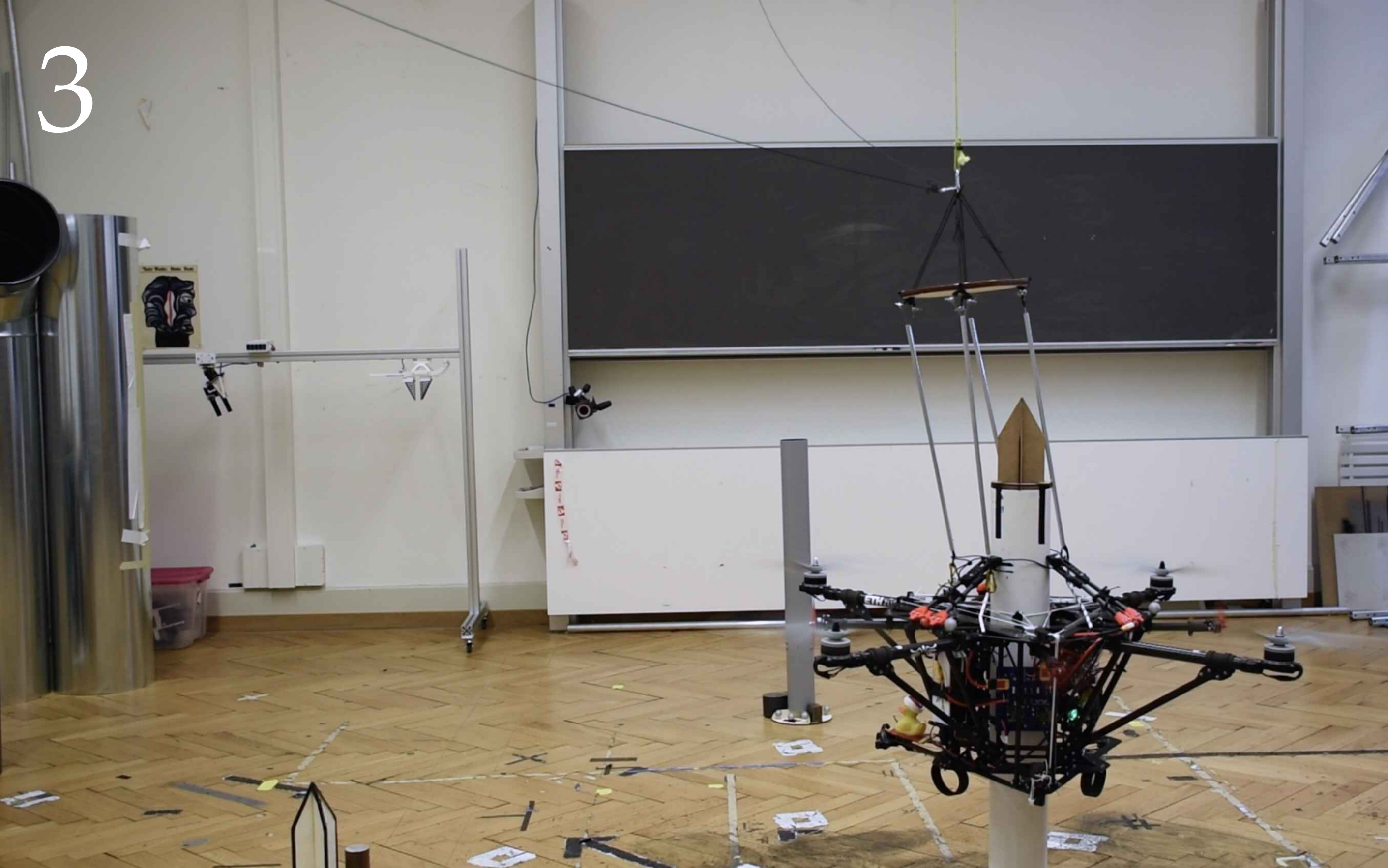}
        \includegraphics[width=.32\linewidth]{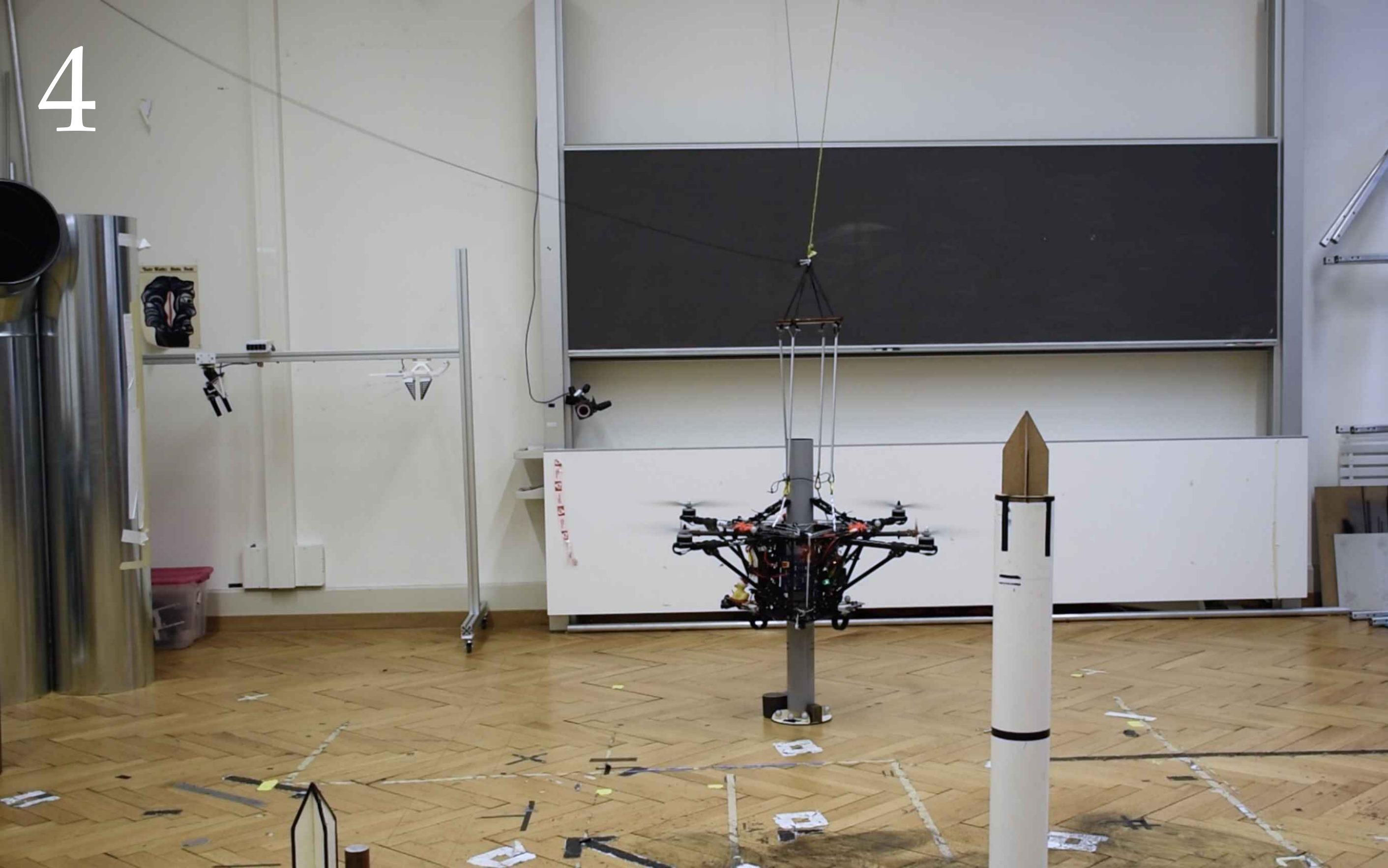}
        \includegraphics[width=.32\linewidth]{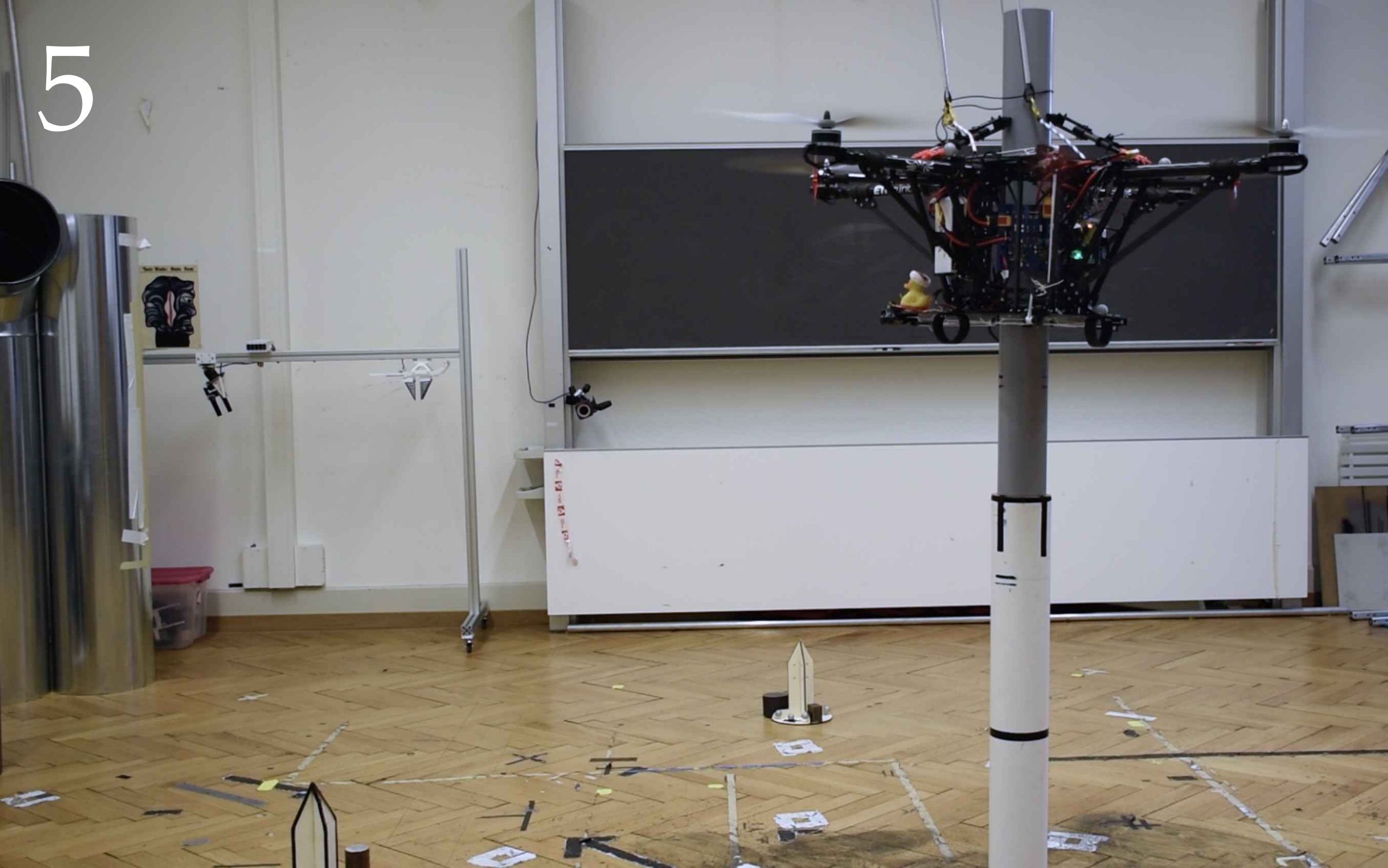}
        \else
        \includegraphics[width=.32\linewidth]{images/submission/eps/7d_image3.eps}
        \includegraphics[width=.32\linewidth]{images/submission/eps/7e_image4.eps}
        \includegraphics[width=.32\linewidth]{images/submission/eps/7f_image5.eps}
        \fi
    \end{minipage}
    
    \begin{minipage}[b]{\textwidth}
        \centering
        \ifarxiv
        \includegraphics[width=.435\linewidth]{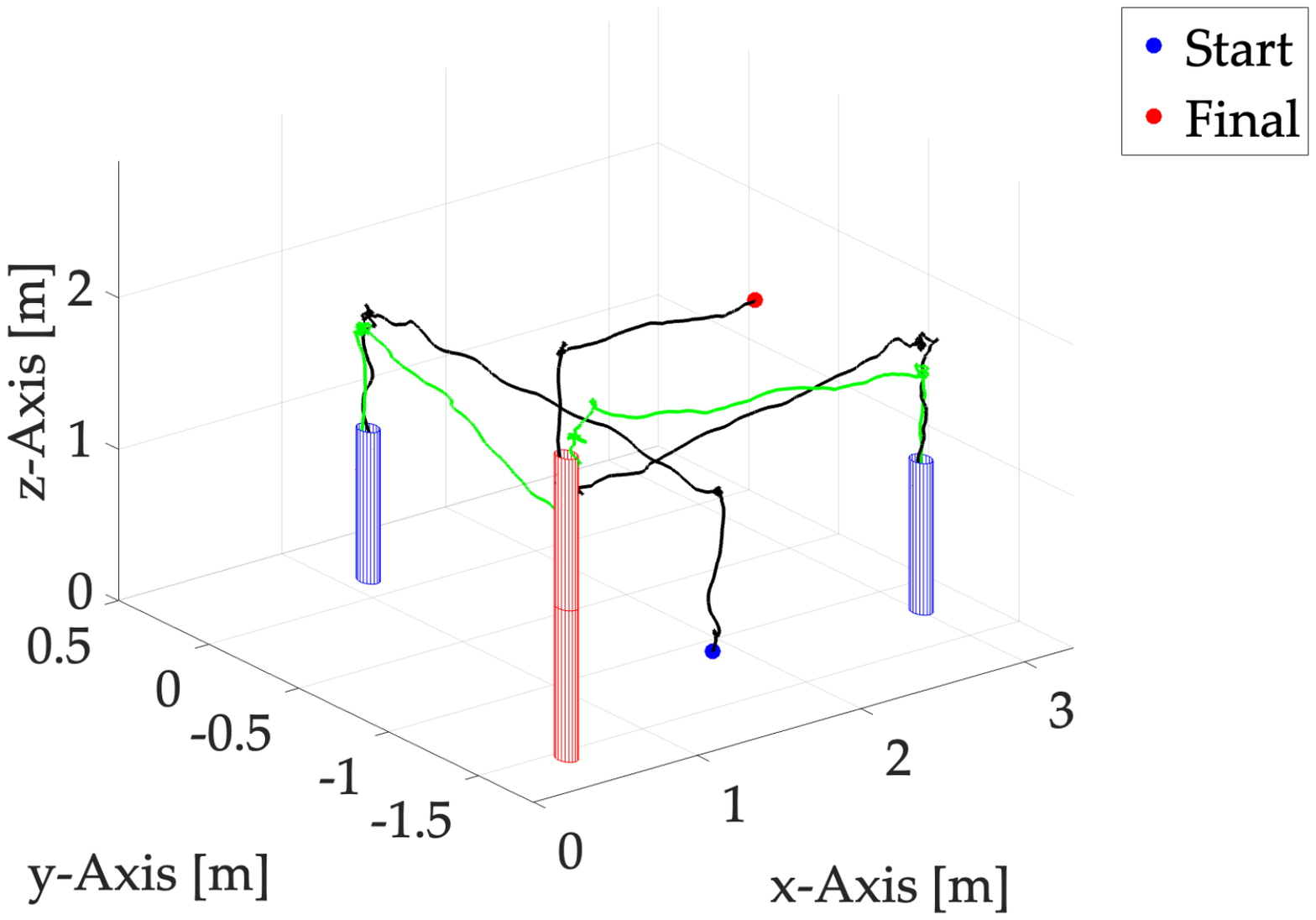}
        \includegraphics[width=.555\linewidth]{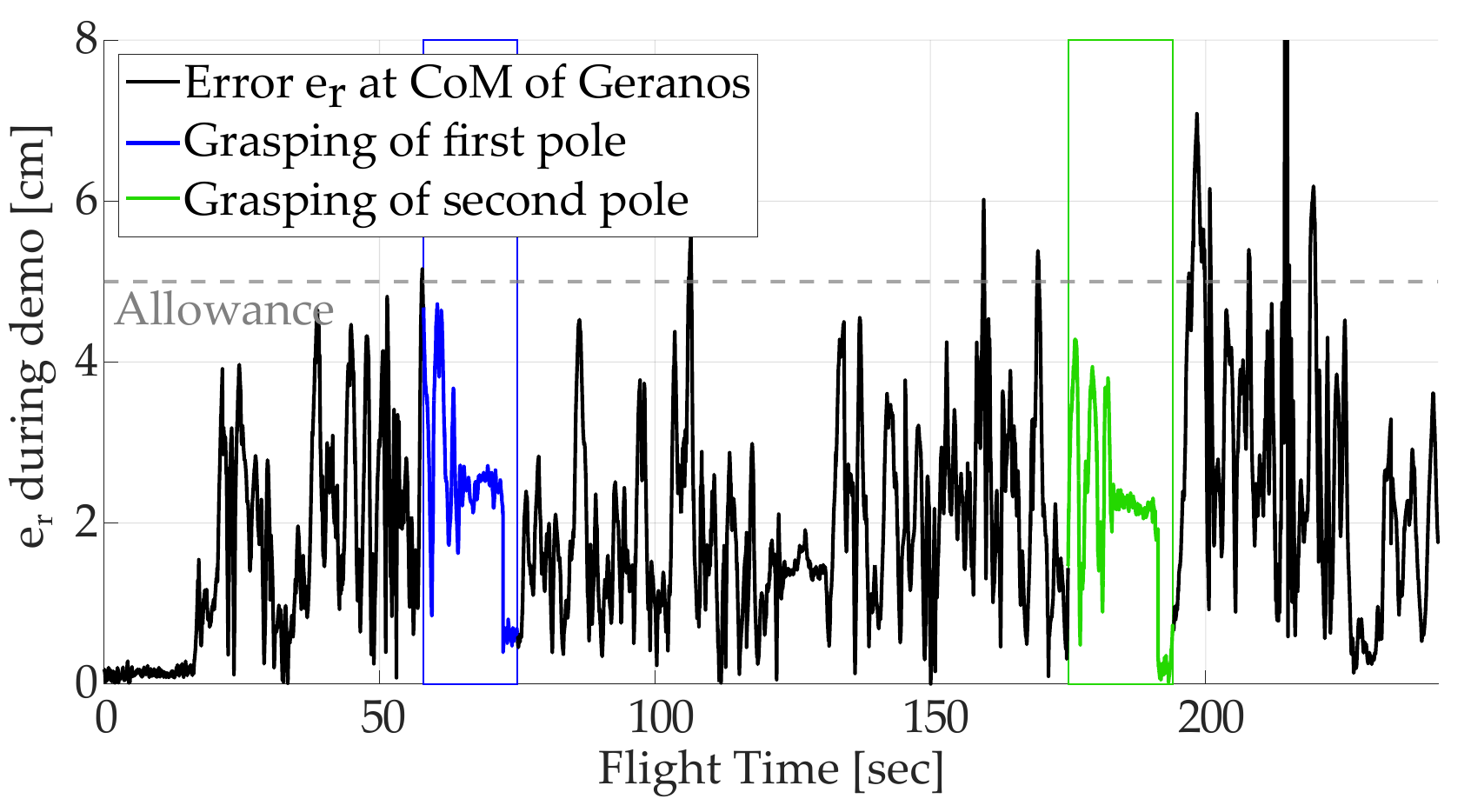}  
        \else
        \includegraphics[width=.435\linewidth]{images/submission/eps/7g_3D_direct.eps}
        \includegraphics[width=.555\linewidth]{images/submission/eps/7h_demo_error_inkscape.eps}
        \fi
    \end{minipage}
    
    \centering
    \caption{\review{Explanation of the course of the demonstration. From top left to bottom right: photo of initial and final pole positions; steps of demonstration; 3D flight path indicating path without payload (black) and with payload and updated controller (green); radial error $e_{r}$ at the CoM \Geranos{} during the entire demonstration. For safety, slack cables are used without disturbing the system.}}
    \label{fig:demo}
\end{figure*}

The demonstration of \Geranos{}, which is used to showcase and evaluate \review{the performance of} our system, can be split into several steps (see \cref{fig:demo}). During the demonstration, two poles are lifted, transported and stacked on top of each other. This illustrates, for instance, \review{how two vertical parts of an aerial lift pylon can be stacked on a relatively small scale.} The poles used in the demonstration are \qty{1}{\m} high and have a weight of up to \qty{2.5}{\kg}. \review{The largest tested payloads that \Geranos{} was able to transport, were poles with a height of \qty{2}{\m} and a weight of \qty{3}{\kg}.\\
The takeoff, flight over the first pole, and descent to the middle of the pole comprise the first part of the procedure (Step 1). We assume that the locations of the poles and mounts are known and set the waypoints in the state machine (see~\cref{fig:controller_structure}) based on this information. 
After positioning itself, a rigid connection between \Geranos{} and the pole is ensured with the help of the gripper. While the gripper is grasping the pole, the controller is updated with the new system's dynamics (total mass and inertia). Step 2 of \cref{fig:demo} shows \Geranos{} lifting the pole and transporting it to its final destination. In step 3, while the pole is lowered and placed into a conical mount, the gripper disconnects the link and the controller switches back to the \Geranos{}' physical dynamics. In steps 4 and 5, \Geranos{} retrieves a second pole and repeats the procedure to stack the second pole on top of the first. This entire procedure takes less than 6 minutes.} \strike{Even though our system is in contact with the environment during grasping and placing the pole, we did not encounter any issues to the flight control.}

\subsection{Precision} 
\label{sec:precision}
\review{
The precision is the most important and quantifiable aspect during assembly operations, particularly during two key stages:
\begin{enumerate*}[label=\roman*)]
    \item \textit{grasping}, when \Geranos{} descends and grasps the pole (steps 1 and 4 in~\cref{fig:demo}); and \item \textit{placement}, when \Geranos{} positions the pole in its final location (steps 3 and 5 in~\cref{fig:demo}).
\end{enumerate*}
 }
\review{
To assess precision, we measure both the position and attitude errors of \Geranos{} and the bottom of the transported pole.
The precision of the vehicle itself is crucial during the grasping stage, while the precision at the bottom of the pole is significant during placement. 
The radial position error of \Geranos{}, $e_r\in\mathbb{R}$, must be kept below the radial tolerance of \qty{5}{\cm}. As stated in \cref{sec:gripper}, this allowance represents the difference between the radius of the pole and the hole of \Geranos{}. The radial position error of \Geranos{}, $e_r$, is calculated as the Euclidean distance between the desired and measured positions in the $x-y$ plane of $\F_w$. Specifically, $e_r = ||\text{diag}(1,1,0)\textbf{e}_p ||$.
 }


\review{
 While $e_r$ can be directly measured, the radial error at the bottom of the pole is determined by propagating the errors at the center of \Geranos{} in position and attitude (as there are no sensors on the pole).
 Under the assumption that the pole is grabbed in the middle, we compute the position error at the tip of the pole}
$\mathbf{e}_{p,tip} \in\mathbb{R}^3$ with: 

\begin{align}
    \mathbf{e}_{p,tip} = \mathbf{e}_p - 
    \frac{L}{2} \mathbf{e}_R^\wedge \, \boldsymbol{z}_{w}.
\end{align}
Here, $L$ \review{represents} the length of the pole, and \review{$\boldsymbol{z}_w$ is the z-axis of $\F_w$}. 
\review{To stack the poles, it is crucial for the radial error ${e}_{r,tip} = ||\text{diag}(1,1,0)\textbf{e}_{p,tip}||$ to be below \qty{5}{\cm}. This tolerance comes from the minimum radius of the pole (stated in~\cref{sec:gripper}) and the geometry of the mounts that the poles are placed on.}

\begin{figure}[t]
    \centering
    \ifarxiv
    \includegraphics[width=\linewidth]{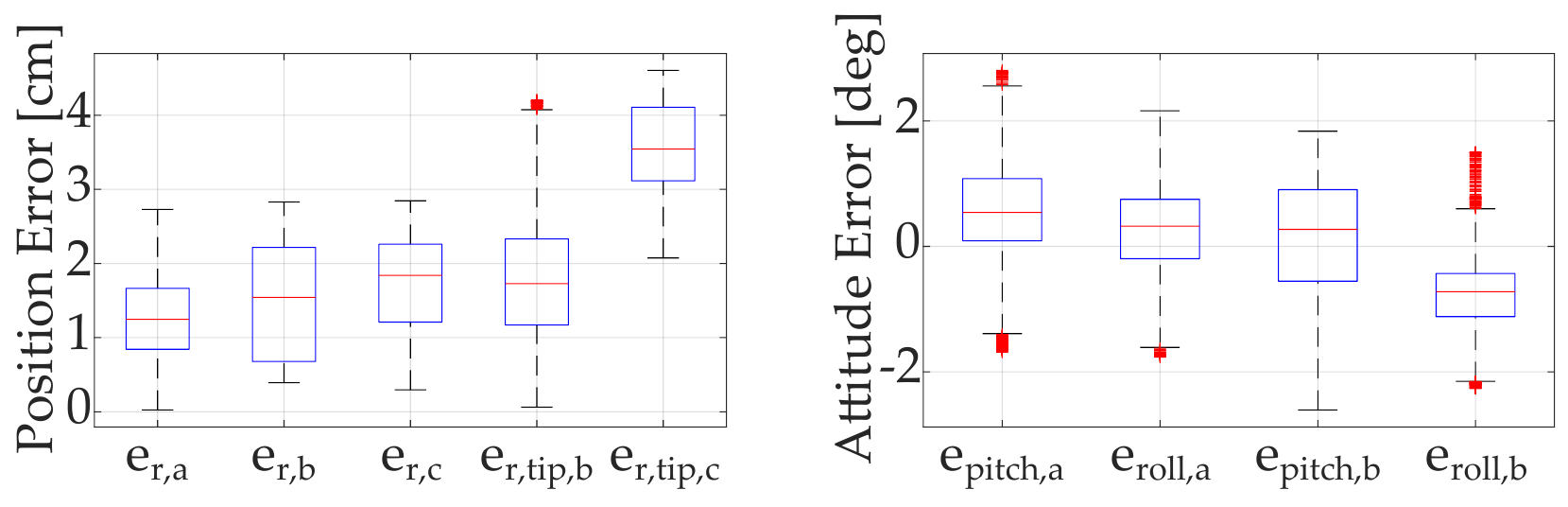}
    \else
    \includegraphics[width=\linewidth]{images/submission/eps/8_boxplots_fig8.eps}
    \fi
    \caption{\review{Left: Radial position error $e_{r}$ at the center of \Geranos{} while hovering (a) without a pole, (b) with a \qty{1}{\m} pole, (c) with a \qty{2}{\m} pole and at the bottom of the pole $e_{r,tip}$ of each of the poles.
    Right: Attitude error in $e_{pitch}$  and $e_{roll}$  of \Geranos{} \review{while hovering} (a) without pole and (b) with a \qty{1}{\m} pole.}}
    \label{fig:boreas_boxplots}
\end{figure}

\review{In \cref{fig:boreas_boxplots}, the radial position error $e_r$ (right) and attitude error (left) are displayed. The radial position error $e_r$ at the CoM of \Geranos{} when hovering without and with a pole up of to \qty{2}{\m} length is less than \qty{3}{\cm}. The error $e_r$ at the bottom of the pole is slightly bigger, especially while transporting a \qty{2}{\m} pole, with a mean up to \qty{3.5}{\cm}. However, these errors are still within the radial tolerance of \qty{5}{\cm} while hovering. The mean attitude errors ${e}_{pitch}$ and ${e}_{roll}$ of \Geranos{} are below \qty{1}{\degree} with maximum attitude errors of less then \qty{3}{\degree}. By observing these errors, we conclude that the accuracy of \Geranos{} in all flight stages is within its allowance of \qty{5}{\cm} and sufficient for the task of grasping and releasing a pole. \\
Additionally, the radial position error $e_r$ at the CoM of the UAV during the entire demonstration is shown in \cref{fig:demo}.
Indeed, $e_r$ is always below the allowance of \qty{5}{\cm} while transporting the poles, with only a few error peaks in free flight.
}



\subsection{
\review{Outdoor Application}
}
\label{sec:outdoor}

\review{While enabling \Geranos{} to perform the assembly task outdoors is of great interest, it is out of the scope of this paper. A full validation of outdoor assembly will be addressed in a future work. However, there are encouraging preliminary results applying visual pose estimation of the poles and position-based visual servoing of \Geranos{} to the task. In combination with GPS real-time kinematic positioning (RTK) for stabilizing the flight outdoors, this has strong potential of enabling \Geranos{} to reliably fulfill the full assembly task presented in~\cref{sec:demo} outdoors.
Preliminary results show that a relative position estimate between \Geranos{} and the payload obtained with a single RGB camera mounted to the bottom plate of the UAV is sufficient for fulfilling the task of grasping a pole with a high success-rate ($>90\%$). That is, in $10$ out of $11$ test flights \Geranos{} was able to successfully approach and grasp the pole. In the remaining test flight the target was lost and the task was aborted as a result. 
The relative position is estimated from the camera's video stream using a keypoint detection network in combination with the Perspective-n-Point algorithm.}

\section{Discussion \& Conclusion}
\label{sec:discussion}

\review{We present a novel UAV, called \Geranos{}, that tackles the challenging problem of transporting and vertically placing poles with an aerial robot.
This research contributes towards removing human intervention in the aerial transportation and vertical assembly of long objects. On a scaled down version of this problem, we demonstrated that the robot has the precision necessary to place and assembly large poles vertically, stacking two poles atop one another.}

\review{The strong performance of the UAV in the proposed task comes down to a number of novelties. 
Firstly, we showed that the lightweight gripping mechanism, composed of a centering mechanism and self locking levers, results in a remarkably rigid grip of the pole. In fact, even when subject to strong accelerations of up to \qty{21}{\meter\per\second\squared}, the relative position and attitude between the payload and \Geranos{} change by no more than \qty{5}{\milli\meter} and \qty{1}{\degree} respectively.
Moreover, the design of the UAV with a hole in its center and additional lateral auxiliary propellers proved to be an effective way to enable \Geranos{} to robustly transport and place a \qty{2}{\meter} long pole with a mass of \qty{3}{\kilogram} and a radius of \qty{5}{\centi\meter} on a conical mount.}

\review{
While we demonstrated that \Geranos{} fulfills the proposed task on its current scale, application to real construction sites requires scaling up. We believe this may be possible with a few modifications. 
For instance, the lifting mechanism would scale quite well given its weight-independent self-locking capabilities. 
However, scaling up brings other challenges like the power consumption. A possible solution could be a fuel powered generator, providing electricity to electrical propeller motors to retain the necessary fast dynamics of the electronic motors. Longer and heavier poles could then be handled by proportionally increasing the size of the UAV.
}

\review{
To extend the design to differently shaped poles (e.g., conical loads) the centering mechanism could be designed differentially, ensuring that the top and bottom strings are constantly tight. Further, the gripper design could use soft materials, providing compliance for irregularly shaped payloads.}

\review{Moving closer to real-world scenarios, future research will extend the system with localization and perception methods. These would reduce its reliance on external positioning systems, allowing it to complete tasks solely with on-board sensors. To demonstrate the viability of such extensions, we outlined preliminary results showing that visual pose estimation and visual servoing can be used to enable \Geranos{} to safely approach and grasp poles without external sensors.
}

\section{Acknowledgements}

In addition to the contributors of this publication, we would like to thank Maximilian Brunner, Severin Laasch, and Paula Wulkop for their contributions to the development of the \Geranos{} system and for their guidance as project coaches.



\bibliographystyle{bibliography/IEEEtran}

\bibliography{bibliography/references}
\addcontentsline{toc}{chapter}{Bibliography}

\vfill

\end{document}